\documentclass[journal]{new-aiaa}
\usepackage[utf8]{inputenc}
\usepackage{textcomp}

\usepackage{graphicx}
\usepackage{epstopdf}
\DeclareGraphicsExtensions{.png,.pdf}
\usepackage{booktabs}
\usepackage{amsmath,amsfonts}
\usepackage{algorithmic}
\usepackage{tabularx}
\usepackage[version=4]{mhchem}
\usepackage{siunitx}
\usepackage{longtable,tabularx}
\usepackage{subcaption}
\usepackage{mwe}
\usepackage{hyperref}
\usepackage{gensymb}
\usepackage[dvipsnames,svgnames,x11names]{xcolor}

\usepackage{fancyhdr}

\title{A Tunnel Gaussian Process Model for Learning Interpretable Flight's Landing Parameters}

\author{Sim Kuan Goh\footnote{Research Fellow, Air Traffic Management Research Institute, NTU, Singapore.}, Zhi Jun Lim\footnote{PhD Candidate, Air Traffic Management Research Institute, NTU, Singapore.}, and~Sameer Alam\footnote{Associate Professor, Air Traffic Management Research Institute, NTU, Singapore.}}
\affil{Air Traffic Management Research Institute, \\School of Mechanical and Aerospace Engineering, \\Nanyang Technological University, Singapore}
\author{Narendra Pratap Singh\footnote{Assistant General Manager (ATM), Airports Authority of India, Lucknow Airport, India.}}
\affil{Airports Authority of India, Lucknow Airport, India.}

\begin{document}

\maketitle
\thispagestyle{fancy}

\begin{abstract}
Approach and landing accidents have resulted in a significant number of hull losses worldwide. Technologies (e.g., instrument landing system) and procedures (e.g., stabilized approach criteria) have been developed to reduce the risks. In this paper, we propose a data-driven method to learn and interpret flight's approach and landing parameters to facilitate comprehensible and actionable insights into flight dynamics. Specifically, we develop two variants of tunnel Gaussian process (TGP) models to elucidate aircraft's approach and landing dynamics using advanced surface movement guidance and control system (A-SMGCS) data, which then indicates the stability of flight. TGP hybridizes the strengths of sparse variational Gaussian process and polar Gaussian process to learn from a large amount of data in cylindrical coordinates. We examine TGP qualitatively and quantitatively by synthesizing three complex trajectory datasets and compared TGP against existing methods on trajectory learning. Empirically, TGP demonstrates superior modeling performance. When applied to operational A-SMGCS data, TGP provides the generative probabilistic description of landing dynamics and interpretable tunnel views of approach and landing parameters. These probabilistic tunnel models can facilitate the analysis of procedure adherence and augment existing aircrew and air traffic controllers’ displays during the approach and landing procedures, enabling necessary corrective actions.

\end{abstract}

\section*{Nomenclature}


{\renewcommand\arraystretch{1.0}
\noindent\begin{longtable*}{@{}l @{\quad=\quad} l@{}}
$GP$ & Gaussian process models\\
$K$ & kernel function defined by $GP$\\
$W$ & $C^2$ - Wendland function\\
$KL$  & Kullback–Leibler divergence \\
$P$ & aircraft's location w.r.t. to ILS signal; $(x, y, l)$\\
$D$ & polar GP coordinate; $(\rho,\theta)$\\
$T$ & tunnel GP (TGP) coordinate; $(\rho,\theta,l)$\\
$(\mu,\sigma)$ & the mean and standard deviation defined by $GP$\\
$(m,S)$ & the mean and co-variance defined by tunnel $GP$\\
$(\mu_t ,\Sigma_t)$ & the mean and co-variance defined in Kalman filter\\
$A$ & transition matrices\\
$C$ & observation matrices\\
$Q$ & transition co-variances\\
$R$ & observation co-variances\\
$\epsilon_{t}$ & white noise\\
\end{longtable*}}

\section{Introduction}

The approach and landing constitute the most safety-critical phase of a flight operation. To execute a safe landing, an aircraft is required to be in a stable configuration, including attitude, airspeed, power/thrust settings as well as runway alignment (touch down point)~\cite{icao2006doc}.
Hence, it is crucial to maintain control of the aircraft and ensure that aircraft parameters comply with specified limits for a stabilized approach and landing~\cite{federal2011airplane}. Failure of compliance can lead to an unstable approach and undesired hazards such as runway excursion and/or a hard landing.

\begin{figure}[ht]
\centering\includegraphics[width=\linewidth]{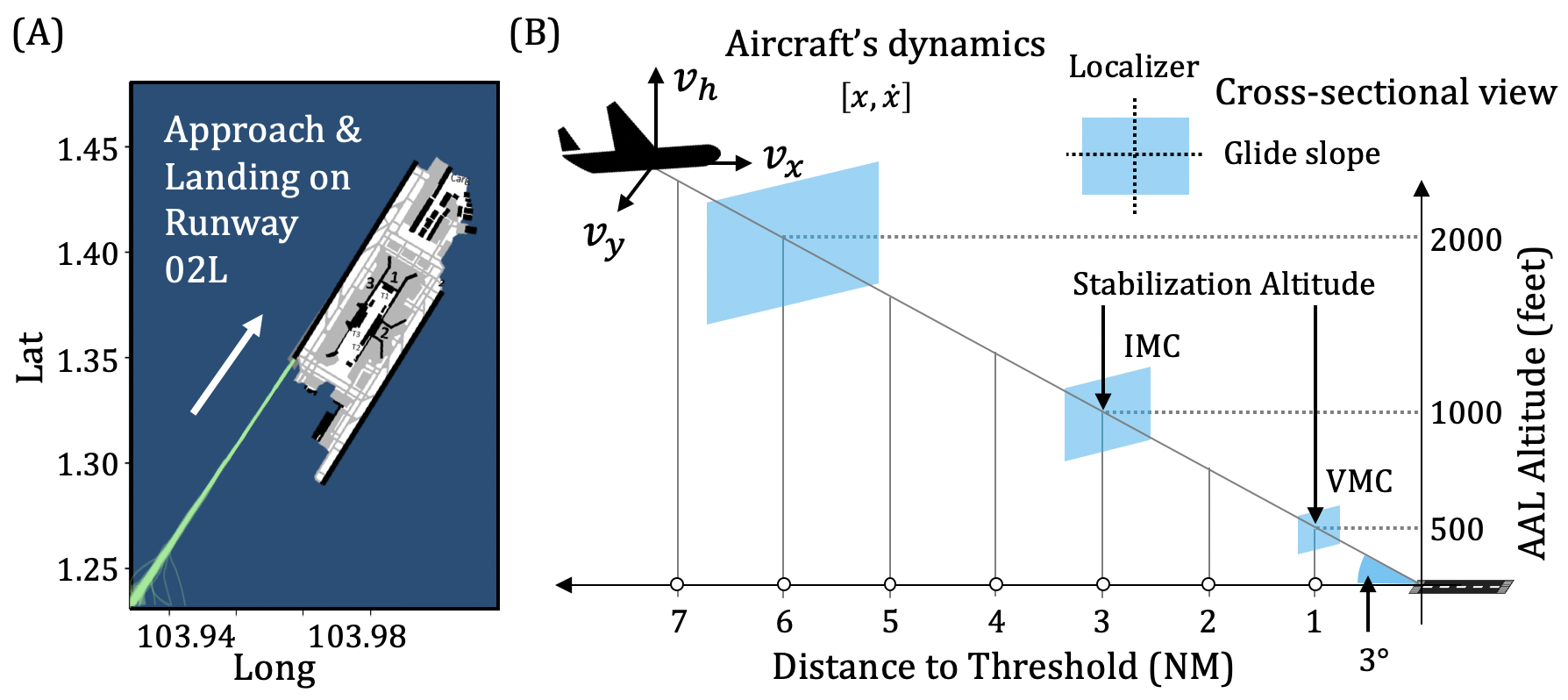}
\caption{Panel (A) shows trajectories of aircraft landing recorded by A-SMGCS. Panel (B) illustrates instrument navigation guidance for approach and landing.}
\label{fig:problem}
\end{figure}

In the Boeing safety report (2007-2016), the final approach and landing constituted $48\%$ of total fatal accidents and on-board fatalities in worldwide commercial jet airplane accidents~\cite{boeing2019statistical}. The report also emphasized unstable approach as a direct or indirect key factor in the majority of the accidents. The international air transport association (IATA), which is an airline regulatory board, showed that failure to perform go-around (aborted landing/ missed approach) was the most predominant factor in runway excursions~\cite{IATA1}. The statistics from the airline industry reveals that an average of $1000$ approaches per day are executed in an unstable configuration. It was found that pilots choose to continue the approach and landing to avoid additional fuel consumption, time and air traffic disruptions. Nevertheless, approximately $30$ of these flights aborted the landing and executed go-around \cite{flight}. 

To mitigate the approach and landing risks, technologies and standard operating procedures (SOP) have been developed and designed. Instrument landing system (ILS), one of the widely used landing aid, facilitates the precision approach via two radio beams by jointly providing both vertical and horizontal landing guidance. While airlines design their own SOP, IATA also has recommended stabilized approach criteria, illustrated in Fig~\ref{fig:problem}, for visual and instrumental meteorological conditions~\cite{IATASOP}.

With the increasing availability and quality of surveillance data, several studies have successfully applied data analytic techniques to investigate the aircraft landing parameters, approach trajectory and management of aircraft energy in the approach and landing phase ~\cite{barratt2018learning,de2014automatic,8547140}. Moreover, these studies derive envelopes of energy during the final approach for a better understanding of energy management in the process of aircraft landing. For example, Yang, Huabo, and Chun~\cite{Yang}, identified the indicated airspeed (IAS) and deviation on localizer to be the most significant factors contributing towards the unstable approaches, using quick access recorder data from B737 aircraft. The study also showed that around 10.3\% of the flights were unstable.

The exponential growth of digital data also takes place in other sectors such as e-commerce~\cite{zhou2019deep}, healthcare~\cite{shilo2020axes,goh2018spatio,goh2016multiway}, etc. In these sectors and industries that embrace data-driven technologies, big data has provided transformative potentials to address a wide range of challenging problems where conventional methods suffer. In addition to big data, the advantages of these data-driven methods are also attributable to the advances of machine learning algorithms and computational power~\cite{domingos2015master,mahajan2018exploring}. Several works have demonstrated the relationship between the enhancing performance of machine learning methods with the increasing amount of data~\cite{mahajan2018exploring,zhou2019deep,shilo2020axes}. By harvesting the knowledge of underlying surveillance data, machine learning techniques have also demonstrated success in many flight operations and air traffic management problems~\cite{lim2020causal,seah2010algorithm,calvo2017conflict,sun2019particle}. Despite achieving state-of-the-art performance, machine learning methods are also widely known for being black-box models, where the underlying mechanism is too intricate for its users to comprehend. The issue has attracted attention from the academic community and several industries to develop interpretable machine learning models~\cite{mueller2019explanation} used in safety-critical environments and settings. In aviation, several interpretable machine learning methods have been used to address issues concerning approach and landing safety~\cite{barratt2018learning,eerland2016modeling, eerland2017gaussian,lim2020causal}.

In this paper, we investigate and develop Gaussian process-based methods to explicitly characterize aircraft's approach and landing parameters. Two variants of tunnel Gaussian process (TGP) models are developed to elucidate aircraft's approach and landing dynamics, leveraging on the data recorded by advanced surface movement guidance and control system (A-SMGCS), which provides surveillance, routing, guidance for the control of vehicles and aircraft (see Section~\ref{asmgcs}, for details). TGP is developed by modifying and hybridizing sparse variational Gaussian process (SVGP)~\cite{hensman2015scalable}, which enables Gaussian process to handle big amount of data, and polar Gaussian process~\cite{padonou2016polar}, which permits cylindrical coordinate modeling. The hybridization allows the proposed TGP to inherit the advantages of the two methods, to characterize and reveal the underlying probabilistic structure of approach and landing trajectory, and to provide interpretable tunnel views of the trajectory.

To assess TGP models, we make a diligent examination of TGP using both synthesized trajectory and recorded trajectory datasets. The synthesized datasets with known probability density functions allow us to qualitatively and quantitatively assess the capability of TGP in reconstructing the structure given trajectory point cloud. Moreover, it facilitates comparative studies against existing methods. Subsequently, TGP is applied to landing trajectories recorded by the A-SMGCS. Based on the reconstructed structure by TGP, we obtain continuously defined and shape varying probabilistic tunnel views of approach and landing profiles as well as parameters defined on it.

The visualization of aircraft, together with the TGP tunnel, can be rendered to ATCOs for the monitoring of stabilized parameters during approach and landing. This visualization augments the predefined rules for a stabilized approach by showing the stochastic nature of the measured trajectories. ATCOs can then take corrective and preventive measures when necessary. By tracking procedure adherence, TGP can inform ATCOs in forward planning of re-sequencing and assist them in monitoring safety procedures of unstable approach and go-around.

This paper is organized as follows: Section~\ref{LR} summarizes the related works, while Section~\ref{method} describes the proposed methodology and framework. In Section~\ref{exp}, we outline the experimental design that includes A-SMGCS data processing, analysis pipeline and evaluation, followed by results and discussion in Section~\ref{rd}, and finally, we present the conclusion in Section~\ref{conclusion}.

\section{Literature Review}\label{LR}

Machine learning methods have been developed to address issues concerning approach and landing safety and performance~\cite{seah2010algorithm,calvo2017conflict,sun2019particle}, which include landing slot allocation~\cite{8835079,7463057}, go-around prediction~\cite{dlh19}, trajectory optimization~\cite{rommel2019gaussian,bonami2013multiphase}, unstable approach detection~\cite{9049174}, and prediction~\cite{sherryimproving}. The functional principal component analysis is used to develop a local anomaly detection algorithm in aircraft landing trajectories by analyzing radar data~\cite{Jarry}. 

Most of these works provide statistical, detective and predictive models for landing analysis. A few studies put forwards to develop supporting tools to aid  procedure design and landing monitoring in real-time. Prominent amongst them uses generative models, such as Gaussian mixture model~\cite{barratt2018learning} and Gaussian process (GP) model ~\cite{eerland2017gaussian}, to generate trajectories similar to historical data. The generated trajectories provide a data-driven approach to examine new procedure design, compared to non-generative methods based on physical equations of motion~\cite{chatterji1999short,chatterji1996route}, trajectory clustering methods~\cite{gariel2011trajectory,li2016anomaly}.   

The aforementioned methods performed time re-sampling, realignment of time $t = 0$, dynamic time warping~\cite{eerland2016modeling, eerland2017gaussian} and trajectory reconstruction~\cite{barratt2018learning} to handle trajectory with variable size or irregular time interval. While these pre-processing steps simplify the analysis, it can distort the temporal information and introduce unnecessary noise. Besides, most of the analyses are discrete, except GP-based methods~\cite{eerland2016modeling, eerland2017gaussian} that model trajectory continuously and provide a "tube" of historical landing profile. GP provides a principal, practical and probabilistic approach to data modeling. GP has demonstrated its practical applications to complex problems ranging from numerical problems such as differential equations~\cite{archambeau2007gaussian}, uncertainty quantification~\cite{bilionis2013multi} to real-world applications such as anomaly detection~\cite{cheng2015video,herlands2018gaussian}, classification~\cite{milios2018dirichlet}, regression~\cite{raissi2019parametric} and optimization~\cite{mlakar2015gp,min2017multiproblem}. In this work, we aim to extend these GP-based studies to model the approach and landing parameters continuously without the aforementioned pre-processing step.

The classical form of GP has been adopted in~\cite{eerland2016modeling, eerland2017gaussian} to model and visualize 3D "tube" of historical aircraft trajectories in Dallas Fort Worth and Denver airport. However, GP suffers from high computational complexity in modeling a large amount of data, which hinders real-time application. Several techniques have been developed to circumvent the complexity~\cite{bauer2016understanding}. In our previous work~\cite{9049174}, we adopted SVGP~\cite{hensman2015scalable} to model aircraft trajectories as well as their dynamics parameters for profile analysis and anomaly detection. SVGP significantly reduces the computational cost of GP and paves the way for real-time detection, compared to offline GP analysis done in~\cite{eerland2016modeling, eerland2017gaussian}. 

Moreover, the "tube" structure derived from historical aircraft trajectories~\cite{eerland2016modeling, eerland2017gaussian} is formed by a union of ellipsoids defined by multivariate Gaussian, which are discrete and symmetrical. Thus, the methods have limited flexibility and cannot handle asymmetrical "tube". For clarity, we will use the word tunnel, which can either be symmetrical or asymmetrical "tube" in the remainder of this paper.

This paper attempts to address the aforementioned issues by modifying the classical GP and developing tunnel GP (TGP) models.  TGP draws inspiration from SVGP~\cite{hensman2015scalable} and polar GP~\cite{padonou2016polar}. While SVGP reduces computational complexity, the polar GP allows the kernel to be defined on polar coordinate and cylindrical coordinates. We develop TGP by hybridizing the strengths of SVGP and polar GP. Therefore, TGP has the advantages of SVGP and polar GP to handle a large amount of data on the cylindrical coordinate, which are apt for 4D trajectories that follow specific paths (i.e., instrumental landing). Cylindrical representations allow explicit modeling of aircraft's trajectory envelop (also known as trajectory dispersion), which is a 2-dimensional manifold that lies in a 3-dimensional space. In this work, cylindrical representation facilitates the explicit characterization of the probabilistic structure of approach and landing trajectories. We build two variants of TGP models for flight's approach and landing parameters and demonstrate that TGP can reveal the underlying probabilistic distribution of flight states (i.e., position and velocity, respectively). Furthermore,  the probabilistic description of TGP provides interpretable tunnel views of flight's approach and landing parameters, which can facilitate comprehensible and actionable insights into flight dynamics for aircrew and air traffic controller.

\section{Methodology}\label{method}

In this section, we first describe a dataset that captures the aircraft approach and landing dynamics, followed by the problem formulation. Subsequently, we describe two related GP methods and discuss how the two methods can be modified to develop two variants of TGP models to learn flight's states (i.e., position and velocity, respectively). The overall architecture of our frameworks is then described in detail.

\subsection{A-SMGCS Data}\label{asmgcs}
A-SMGCS~\cite{854020} provides ground-based surveillance for routing and guidance of the aircraft during ground movement operations, which is particularly advantageous and sometimes one of the mandatory requirements for airport operation in low visibility conditions. The system captures the arrivals and departures as well as surface vehicles at airports. Besides supporting ground movement operations, the system can be used for tracking and recording aircraft movements in the vicinity of airports.

The A-SMGCS measurements comprise the positional information of aircraft in UTM WGS84 format (latitude, longitude, altitude), aircraft identity, velocity, and aircraft wake-turbulence category. These measurements are recorded with a high temporal resolution at a sampling rate of 1 Hz. 
These essential features can provide insight into the instrument landing dynamics. One month of A-SMGCS data collected from Singapore Changi airport is used in this study.



\subsection{Problem Formulation}

Current procedures for stabilized instrument landing are mainly described using a set of rules on landing parameters (e.g., airspeed, sink rate, heading), where the judgment of go-around is based on compliance with these rules. In this paper, we attempt to aid the decision-making process by devising continuous probabilistic values of compliance using data-driven models. Given historical A-SMGCS data, the aircraft's dynamics during a precision runway approach can be modeled. Building on our previous work~\cite{9049174} that develops probabilistic bounds on lateral, vertical, speed and track angle landing profiles for anomaly analysis and detection, we aim to take a step forward from side views (i.e., lateral and vertical) analysis to probabilistic tunnel view analysis. 

In A-SMGCS data, the positional data is similar to the point cloud studied in light detection and ranging (Lidar)-based computer vision~\cite{dandois2010remote}. However, Lidar data generally lies on a surface. In contrast, the positional data is volumic, dense, and contains speed. In the aviation context, the positional data encapsulates landing motion aided by ILS, which has high regularity. The research problem is to reconstruct the probabilistic structure given a volumic point cloud of the landing trajectory, which can reveal the variation of approach and landing parameters during instrumental landing. Specifically, we characterize the underlying probabilistic structure explicitly in cylindrical coordinates $(\rho,\theta,l)$ where $(\rho,\theta)$ is polar coordinate defined on the longitudinal axis along the pole in cylindrical coordinates.

\begin{figure}[ht!]
    \centering
    \includegraphics[width=\linewidth]{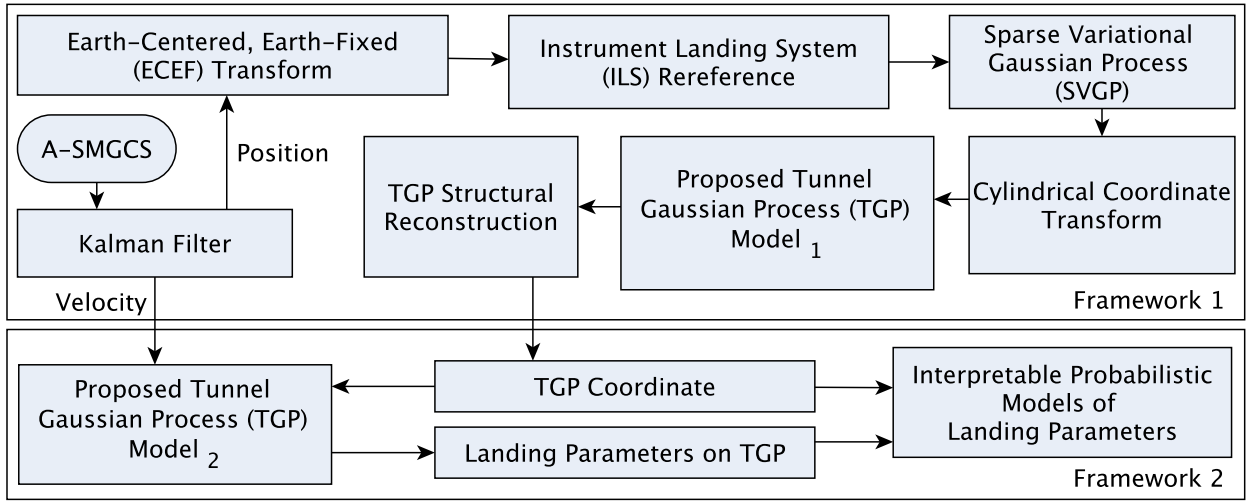}
    \caption{Our proposed TGP frameworks to learn interpretable approach and landing parameters.}
    \label{fig:Flowchart}
\end{figure}

The overall architecture of our proposed frameworks is shown in Fig.{~\ref{fig:Flowchart}}, which comprises two main frameworks to learn a trajectory dataset with position and velocity information to construct the interpretable probabilistic models of landing parameters. The first framework comprises procedures such as data preparation (that involves earth-centered earth-fixed (ECEF) transformation, Kalman filtering, extraction of localizer and glide slope deviation, normalization), estimation of the non-linear pole in cylindrical coordinate using SVGP, cylindrical coordinate transformation. Given the preprocessed data, TGP models 1 processes the position data by combining SVGP (for reducing the computational complexity) and polar GP (for dispersion modelling in cylindrical coordinate). TGP model 2, from the second framework, learns the landing parameters (i.e., velocity) that are defined on TGP model 1.

\subsection{Sparse Variational Gaussian Process}
Given $D = {(x_i,y_i)_{i=1}^{n}} = (X,y)$, where $X$ is $d$ dimension input${x_i}_{i=1}^n \in R^d$ and $y$ is output ${y_i}_{i=1}^n \in R$, and $n$ is the total number of data instances. Gaussian process model learns the probabilistic mapping function $f(x)$:

\begin{equation}
y_i = f(x_i) + \epsilon_i
\label{eqn:firstgp}
\end{equation}

\noindent where we model $f$ using Gaussian process $GP(m,K)$ with mean function $m$ and covariance function $K$ and $\epsilon_i$ is Gaussian white noise with $N(0,\sigma^2)$. The $x_i$ can be time-based or space-based indexing, depending on problems. As timestamp is discrete in surveillance data, continuous space-based indexing is selected to learn the approach and landing dynamics, which are inherently continuous.

The predictive distribution for a new data point $x_*$ can be obtained as follows:

\begin{equation}
p(y_*|x_*,X,y) = N(\mu_*,\sigma_{*}^{2})
\label{eqn:gp}
\end{equation}

\noindent where $\mu_* = K_{*n}(K_n+\sigma^2I)^{-1}y$ and $\sigma_*^2 = K_{**}-K_{*n}(K_n+\sigma^2I)^{-1}K_{n*}+\sigma^2$. $K_n$ is the covariance matrix between training data, $K_{**}$ is the covariance matrix between the new data points, $K_{*n}$ is the covariance matrix between the new data points and training data and $K_{n*}$ is the covariance matrix between the training data and the new data points. Hence, the GP provides prediction $\mu_*$ and uncertainty $\sigma_*$ given $x_*$. One issue of the basic form of GP is the computational complexities of $O(n^3)$ and storage complexity of $O(n^2)$. The complexities hinder the application of GP to big data.

Sparse Variational Gaussian Process (SVGP) was proposed to circumvent the computational and storage complexity of the basic form of GP, paving the way to model millions of data points~\cite{hensman2015scalable}. To reduce the complexity, SVGP combines stochastic variational inference with inducing variables for GP training. SVGP introduces input-output pairs $Z$, $u$ as inducing variables, where $u$ of the size $M$ that contains values of the function $f$ at the points $Z$ that lies in the same input data space of $x$. The variable $u$ works as a global variable with a variational distribution $q(u) = \mathcal{N}(u \mid m_u, S)$, a multivariate normal distribution with mean $m_u$ and covariance $S$. SVGP optimizes the variational bound $\mathcal{L}$ as follows:

\begin{equation}
\begin{split}
\mathcal{L} &= \sum_{i=1}^{n} \Biggl\{ log \mathcal{N}(y_i \mid k_i^\top K_{mm}^{-1}m_u,\beta^{-1}) \\
 &\quad - \frac{1}{2} \beta \tilde{k}_{i,i} - \frac{1}{2} tr(S\Lambda_i) \Biggr\} \\
 &\quad - KL(q(u) \parallel p(u))
\end{split}
\end{equation}

\noindent where $k_i$ is the $i^{th}$ column of $K_{mn}$ and $\Lambda_i = \beta K_{mm}^{-1} k_i k_i^\top K_{mm}^{-1}$, $K_{mm}$ and $K_{mn}$ are kernel between inducing points and between inducing points and training data points, considering independent Gaussian noise of precision $\beta$. KL is the abbreviation for Kullback–Leibler divergence. $\tilde{k}_{i,i}$ is the $i^{th}$ diagonal element of $\tilde{K} = K_{nn} - K_{nm}K_{mm}^{-1}K_{mn}$. The derivation and natural gradients to optimize the bound are provided in~\cite{hensman2015scalable}. SVGP reduces the complexity from $O(N^3)$ to $O(NM^2)$ and enable stochastic optimization. The complexity of inference is $O(M^2)$.

However, SVGP's paper~\cite{hensman2015scalable} does not investigate polar and cylindrical coordinates that involve geodesic distance, which causes the kernel to be non-positive definite.

\subsection{Polar Gaussian Process}
Polar GP~\cite{padonou2016polar} is developed for prediction on circular domains by exploiting the geometry of a disk. It is designed to address two physical processes involving rotation and diffusion from the center of a disk in microelectronics and environmental engineering, where geodesic distance is crucial. A unit disk $D$ can be defined as:

\begin{equation}
D = \{ ( \rho cos \theta,\rho sin \theta),\rho \in [0,1],\theta \in [-\pi,\pi] \}
\end{equation}

\noindent where $\rho$ denotes radial coordinate, $\theta$ denotes angular coordinate. $D$ can be extended to cylinder $C$ by $[0,1] \times D$. In~\cite{padonou2016polar}, two sufficient conditions for positive definiteness of kernel are described. $C^2$ - Wendland function, a compactly supported radial basis function (rbf) on $\mathbb{R}$, guaranteed the kernel on the geodesic distance to be positive definite. The function is as follows:

\begin{equation}
W_c(t) = \left ( 1+\tau \frac{t}{c}  \right ) {\left ( 1- \frac{t}{c}  \right )}_{+}^{\tau}
\label{eqn:wendland}
\end{equation}

\noindent where $t$ is the geodesic distance, $\{{c \in \mathbb{R}}, 0 < c \leq \pi \}$ and $\{ \tau \in \mathbb{R}, \tau \leq 4 \}$. $(\dot)_+$ is a rectifier function that is linear to a positive input but zeroes to negative input. For the case of a disk, $c = \pi$.

While polar GP works can perform prediction on circular domains, it is not directly applicable to A-SMGCS positional data, which is non-concentric. Besides, it does not address structural reconstruction given the data.

\subsection{Tunnel Gaussian Process}
Here, we discuss two variants of the proposed TGP models, describing how we combine the strengths of SVGP and polar GP by addressing some of their limitations. TGP model 1 is built to reconstruct the probabilistic tunnel of approach and landing, while TGP model 2 learns other landing parameters (e.g., velocity).

In polar GP~\cite{padonou2016polar}, geodesic distance $d_g$ is defined as follows:
\begin{equation}
d_g(\theta,\theta') = acos (cos(\theta-\theta'))
\end{equation}

\noindent However, the derivative of this function is not continuous everywhere, specifically at $\theta-\theta' = 0$ or $\pi$. This hinders any gradient-based optimization methods. Moreover, the trigonometric operation has a higher computational cost compared to normal operation. We modify it to $d_g(\theta,\theta') = min(abs(\theta-\theta'),2\pi - (\theta-\theta'))$ and provide the derivative of $d'_g$:

\begin{equation}
d'_g(\theta,\theta') = - sign(\theta-\theta') \times sign(abs(\theta-\theta')-\pi)
\end{equation}

Moreover, we also modify the polar GP to handle non-concentric shapes by changing the radial coordinate to the diametral coordinate for $y$ in Eqn.~\ref{eqn:firstgp}. Data points that lie in the same diameter defined by an angle $\theta$ are symmetrical to mean $\mu_*$ and has a boundary defined by $\sigma_*$, where $\mu_*$ and $\sigma_*$ are learned from data, defined in Eqn.~\ref{eqn:gp}. Thus, it is flexible to handle non-concentric data. The treatment of diametral coordinate can be incorporated directly in SVGP; hence, no transformation is required to the data defined on polar coordinates. We modify the $z$, $m$, $S$ in SVGP to achieve diametral coordinate by exploiting the symmetrical structure of a diameter, which are formed by two symmetrical radiuses. The $u$ in $z$ and $m$ are augmented with an additional $u'$ and $m'$, which are rotation of $u$ and $m$ by $\pi$. $S$ is augmented to be $S_d$ as follows:

\begin{equation}
S_d = 
\begin{bmatrix}
S & S'^\top \\ 
S' & S
\end{bmatrix}
\end{equation}

\noindent Besides, the lower triangular property of $S$ and $S'$ can be exploited to reduce the number of training parameters. To guarantee positive definiteness, Eqn.~\ref{eqn:wendland} with $c = \pi/2$ was chosen for the symmetrical property of a diameter. This modification allows TGP to handle non-concentric data. 

GP assumes the parameter $\sigma$ in the likelihood to be a constant. In many applications, the likelihood is required to be input dependent, where $\epsilon_i \sim N(\cdot,\sigma_i^2)$. Based on the observation in our previous work~\cite{9049174}, the variation of aircraft dynamics is observed to be changing along with the landing phase. The phenomenon, known as heteroscedasticity in statistics, is also reported in{~\cite{kersting2007most}}. In order to model the variation during the approaching phase, heteroscedastic likelihood{~\cite{kersting2007most}}, which can capture the non-constant variation, has to be incorporated into the SVGP and our TGP model. The likelihood allows SVGP and our TGP to flexibly model the uncertainty bounds (i.e., deviation in parameter) for different parts of the approach and landing phase.
 
TGP model 1 is a process of two components: angular process $A(\theta)$ and longitudinal process along with the ILS path $L(l)$. The kernel function $k$ is defined by tensor product:
 
\begin{equation}
k(u,u')= k_a(\theta,\theta')k_l(l,l')
\end{equation}

\noindent where $u = (\theta,l)$, $k_l$ is longitudinal kernel along the ILS path. 'rbf' kernel is used for $k_l$. $k_a$ is an angular kernel where $C^2$ - Wendland function is used.

For landing parameters ($S_x$, $S_y$, $S_h$ for longitudinal speed, lateral speed and descent rate) defined with respect to ILS, $y=S$ in Eqn.~\ref{eqn:firstgp}, we can include the radius process $R(\rho)$ for TGP model 2. The kernel function $k$ is defined by tensor product:
 
\begin{equation}
k(u,u')= k_r(\rho,\rho')k_a(\theta,\theta')k_l(l,l')
\end{equation}

\noindent where $u = (\rho,\theta,l)$, $k_r$ is radius kernel where 'rbf' kernel is used. $k_a$ is angular kernel where $C^2$ - Wendland function is used, with $c = \pi$. $k_l$ is longitudinal kernel along the ILS path.

\section{Experiments}\label{exp}
Prior to applying TGP to learn A-SMGCS data, we empirically examine TGP on synthesized trajectories, which have continuously varying shapes and known probability density functions, as shown in Table{~\ref{table:Syndata}}. Multivariate Gaussian and 5-petalled rose are used as building blocks to create structural variation. Trajectories with three types of probabilistic structures are defined to assess TGP. Based on the probabilistic structures, a dense point cloud can be generated by sampling data points in the structures. TGP is examined based on its capability in reconstructing the probability density functions given dense point clouds. We also compare TGP against existing methods for trajectory learning. Subsequently, TGP can be applied to A-SMGCS data to reconstruct the probabilistic tunnel views of the approach and landing profile as well as the speed defined on the tunnels.

\subsection{Synthesized Data}

\begin{table}[]
\renewcommand*{\arraystretch}{1.4}
\begin{tabularx}{\textwidth}{lXXX}
\toprule
& \textbf{$S_1$} & \textbf{$S_2$} & \textbf{$S_3$} \\ \midrule
Description & Gaussian distribution on non-linear pole &  Twisted 5-petalled rose & Gaussian mixtures\\ \midrule
Pole $p(l)$ &  $(x_p,y_p) = (2l^2,2l^2)$ & $(x_p,y_p) = (0,0)$ & $(x_p,y_p) = (0,0)$ \\ \midrule
2D structure & $Gaussian(\begin{bmatrix} 0 \\ 0 \end{bmatrix},\begin{bmatrix}
{\sigma}_a^2 & 0\\ 0 & {\sigma}_b^2
\end{bmatrix})$ 
& $r(\theta,l) = \alpha (3+ cos(5(\theta+\pi \phi/5))) $   
&  $Gaussian(\begin{bmatrix} 0 \\ 0 \end{bmatrix},\begin{bmatrix}
\sigma_a^2 & 0\\ 
0 & \sigma_b^2
\end{bmatrix})$ + $Gaussian(\begin{bmatrix}
0
\\ 0
\end{bmatrix},\begin{bmatrix}
\sigma_b^2 & 0\\ 
0 & \sigma_a^2
\end{bmatrix})$ \\ \midrule
Parameter variation & $l \in [-1,1], \sigma_a = 2 - l, \sigma_b = 2 + l $ & $l \in [-1,1], \phi = 2\pi l/5, \alpha = 2 - |l| $ &  $l \in [-1,1], \sigma_a = 3, \sigma_b = 1$\\ \midrule
3D structure & 
    \includegraphics[width=0.78\textwidth]{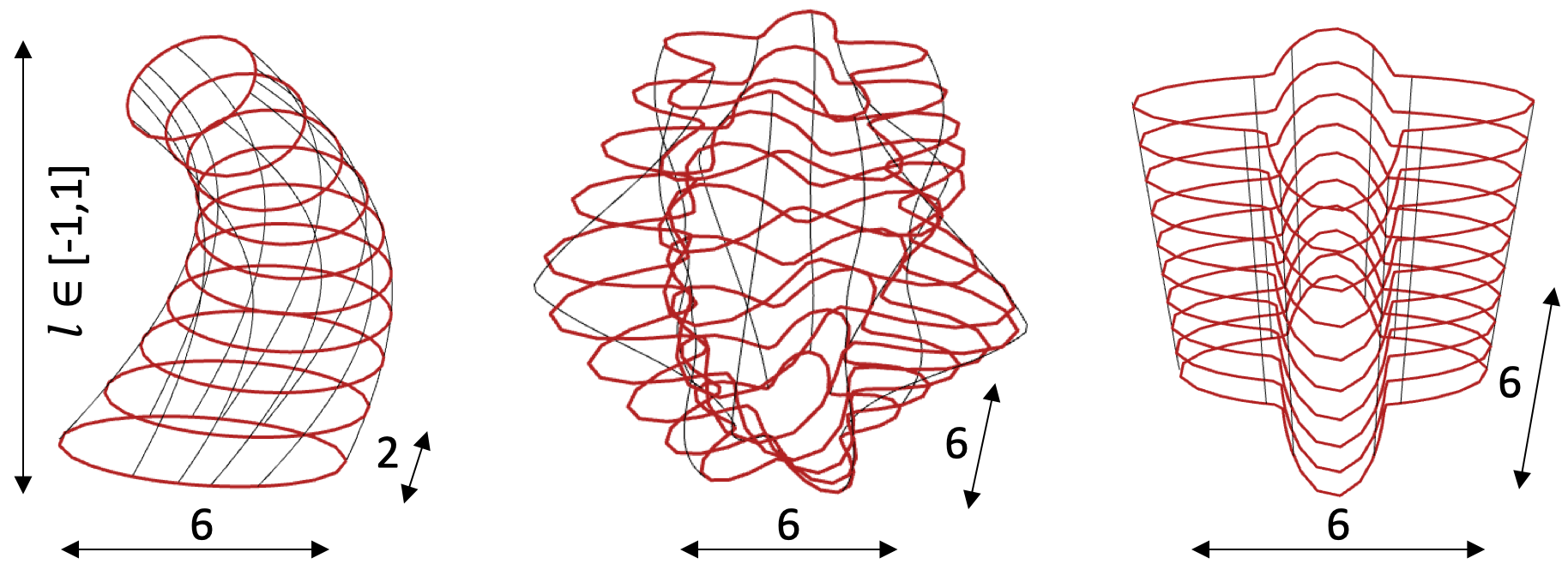}
      \\ \bottomrule
\end{tabularx}
\caption{Synthesized dataset (drawing not to scale). The wireframes show the skeletal outline of the 3D structure, horizontally (brown curves, thick) and vertically (black curves, thin).}
\label{table:Syndata}
\end{table}

Table.~\ref{table:Syndata} illustrates the generating process of the synthesized datasets, the ground truth structure and the dense point cloud. The structures have characteristics such as concentric, non-concentric, symmetric, asymmetric and size variation. To obtain the synthesized datasets, three structures are defined, which comprise pole function $p(l) = (x_p,y_p)$ and radius function $r(p,l,\theta)$. $l$ denotes the axis along the pole in cylindrical coordinates in the $[-1,1]$. Given $l$, the $p$ or center of cylindrical coordinate can be defined. A cross-sectional surface at $l$ can be defined with $\rho$ and $\theta$:

\begin{equation}
    S_i(l,\theta) = (x_s,y_s) = (x_p+\rho cos(\theta),y_p+\rho sin(\theta))
\end{equation}

\subsection{Operational Environment}
Based on the published ILS landing procedure for runway 02L of Singapore Changi airport, an aircraft must horizontally align with ILS while maintaining a track angle of $23 \degree$ and vertically descending on glide slope of $3 \degree$ from a fix ANUMA, shown in Fig~\ref{fig:problem}. The permissible horizontal and vertical deviations on ILS are available to the pilots and they monitor the adherence to the ILS. The track data is collected using A-SMGCS, with a total of 2212 medium wake category aircraft and 2305 heavy wake category aircraft from runway 02L.

\subsection{Experimental Pipeline}
The experimental steps include data preparation, feeding of data into the TGP pipeline, model evaluation and case studies on A-SMGCS data.

\subsubsection{Data Preparation}
While synthesized data defined in Section~\ref{exp} is ready for experiments, A-SMGCS requires additional processing steps. The pre-processing of A-SMGCS data includes ECEF and ILS re-reference discussion in Section~\ref{method}. Given positional data $(x,y,z)$, we represent the data as $(x,y,l)$ where $l$ is the longitudinal axis along the pole in the coordinate system.

\subsubsection{ECEF and ILS Deviation}
We adopt the ECEF coordinate system{~\cite{9049174}}, which is also known as the "conventional terrestrial" coordinate system. In this coordinate system, the coordinates of trajectories $(x,y,z)$ share the same unit. The unit vector $\overrightarrow{a}$ in ECEF, which points from the final approach fix $(x_{FAF},y_{FAF},z_{FAF})$ to the runway threshold, provides the longitudinal axis along the glide slope. We defined a vector $\overrightarrow{b}$ in ECEF that originates from the final approach fix pointing vertically downward to the ground. By computing the cross product of $\overrightarrow{b}$ and $\overrightarrow{a}$, we can obtain a unit vector $\overrightarrow{c}$ that represents lateral deviation from the localizer. Subsequently, the cross product between $\overrightarrow{c}$ and $\overrightarrow{a}$ gives $\overrightarrow{d}$, which represents the deviation from glide slope. We reference the trajectory to the given ILS path and obtain $P = (x_{lateral},x_{glide slope},l)$, which is the dot product between $(x - x_{FAF},y-y_{FAF},z-z_{FAF})$ and $(\overrightarrow{c},\overrightarrow{d},\overrightarrow{a})$.

\subsubsection{Kalman Filter}
Kalman filter (KF), an unsupervised algorithm for object tracking in a continuous state space, is a widely used state estimation technique and has demonstrated wide-ranging engineering applications. Assuming a linear time-invariant dynamical system, system' states and measurements can be defined as:

\begin{equation}
\begin{split}
&x_0 \sim N(\mu_0 ,\Sigma_0)\\
&x_{t+1} = A_tx_t+b_t+\epsilon_{t+1}^1\\
&y_t = C_tx_t+d_t +\epsilon_{t+1}^2\\
&\epsilon_{t}^1 \sim N(0 ,Q)\\
&\epsilon_{t}^2 \sim N(0 ,R)
\end{split}
\end{equation}

\noindent where $x_t$ is system state at time $t$, $y_t$ is measurement at time $t$, $Q$ and $R$ are transition and observation co-variances, $A$ and $C$ are transition and observation matrices, $b_t$ and $d_t$ are transition and observation offsets. The parameters of KF can be estimated using the expectation maximization (EM) algorithm~\cite{ghahramani1996parameter}. KF is adopted to estimate the descent rate of aircraft, which is not available in the A-SMGCS we used.

\subsubsection{Cylindrical Coordinate Transformation}
Prior to the cylindrical coordinate transform, it is crucial to ensure that the pole of the cylindrical coordinate passes through the center of trajectories. Based on the observation, 4D trajectories are not linear. To address this, the continuous mean in lateral $x_p(l)$ and glide slope $y_p(l)$ deviation of our previous work~\cite{9049174} by SVGP can be used to model a non-linear pole $P(l) = (x_p,y_p,l)$.  

Subsequently, we can transform the coordinate $(x_{lateral}-x_p,x_{glide slope}-y_p,l)$ to TGP coordinate $T = (\rho,\theta,l)$.

\subsubsection{TGP Pipeline}
The pipeline of our TGP framework is the same for synthesized data and pre-processed A-SMGCS data. First, the standard scaler is applied to $x$ and $y$ to normalize them, while the min-max scaler is used for $l$ such that it ranges between -1 and 1. Next, SVGP is used to derive the non-linear pole for cylindrical coordinate transformation. After the transformation, TGP is trained to model the structure of the dense point cloud. 

\subsubsection{Hyper-parameters}
For synthesized data, we set $M$ to be 10 for SVGP, and varies $M$ for TGP according to the $R_{TGP}$, which is the size of $(\mu,\sigma)$ pairs used, by selecting the closest integer to the $R_{GMM}$ of GMM-based methods, which have an increasing number of multivariate mixture components ${1,2,4,8,16}$. The selected values can be found in Fig~\ref{fig:compare}. The training batch size is 400 for SVGP and 800 for TGP.

For A-SMGCS data, the $M$ for SVGP is 100 and it is 300 for TGP. The training batch size is 400 for SVGP and 800 for TGP. For speed defined on TGP bounds, we select $M$ to be $8^3$ and training batch size to be 1000. These parameters are empirically chosen to achieve the main objective of this work. Higher $M$ and training batch size can always be selected for greater modeling resolution. However, it will also result in higher computational and storage costs. All the other parameters are optimized using stochastic gradient descent. Adam optimizer~\cite{kingma2014adam} is used in this work.

\subsubsection{Evaluation}
TGP is assessed using both synthesized data and real-world landing parameters from A-SMGCS. Given synthesized data and ground truth, TGP can be assessed both qualitatively by visual inspection and quantitatively by error metrics (e.g., root-mean-square error). $5x10^5$ data points are synthesized.

We also compared the TGP against two variants of Gaussian mixtures models (GMM), which are probabilistic models that can capture the underlying data distribution using a finite mixture of multivariate Gaussian distributions. We compare TGP with the original form of GMM and the modified GMM (mGMM) proposed in{~\cite{barratt2018learning}} for trajectory learning. As GMM models are discrete models, it could not handle trajectory data with variable intervals and variable numbers of data points. Usually, interpolation is required to align the data. Here, we simplify the problems for GMM and mGMM by providing perfect interpolation at 10 evenly separated planes along the pole of the synthesized structures. However, our TGP models are trained on the original point clouds, which have both variable intervals and variable numbers of data points. 

Given the probability density functions (PDF) approximated by GMM, mGMM, and TGP, we can compute the root mean square error (RMSE) using the ground truth PDF. The RMSE is computed on a regular grid of width 0.2 at 10 evenly separated planes along the pole of the synthesized structures. The range of evaluation for $S_1$ is $[-10, 15]\times[-10, 15]$, while the range for $S_2$ and $S_3$ is $[-10, 10]\times[-10, 10]$.

\section{Results and Discussion}\label{rd}
TGP is empirically assessed using synthesized datasets and flight's landing parameters. Synthesized data provides ground truth for model assessment, while the landing parameters demonstrate the practical application of the model in real-world data.

\subsection{Synthesized Datasets}

\begin{figure}[ht!]
    \centering
    \includegraphics[width=\linewidth]{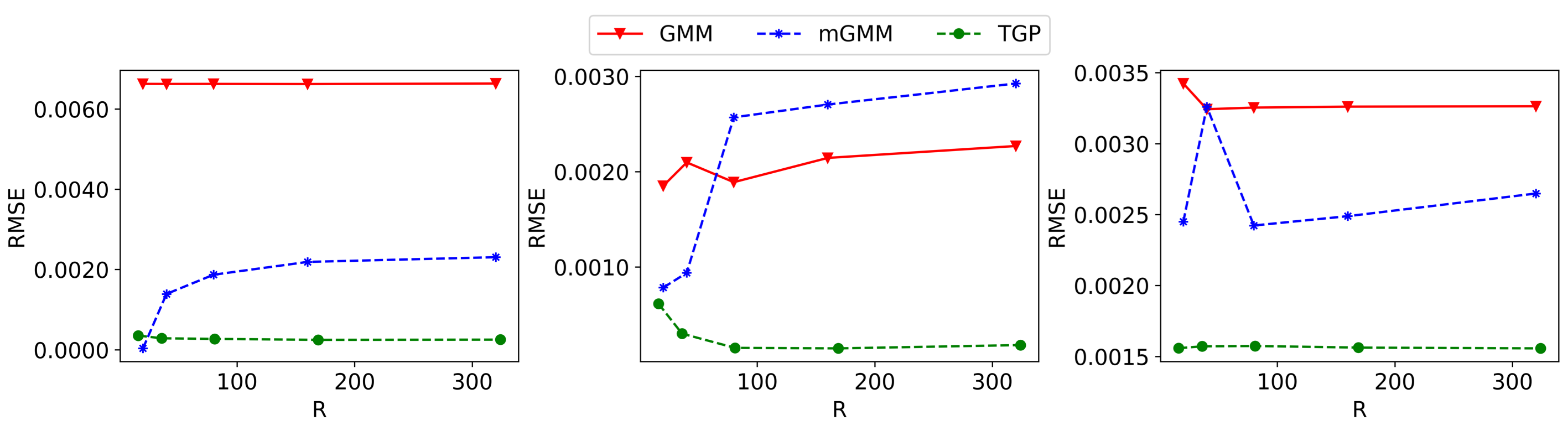}
    \caption{RMSE between the actual PDF and the approximated PDF using GMM, mGMM and TGP, for structure $S_1$ (left), $S_2$ (middle) and $S_3$ (right). $R$ is the size of $(\mu,\sigma)$ pairs in each algorithm.}
    \label{fig:compare}
\end{figure}

The RMSE results of GMM, mGMM and TGP, in reconstructing the PDF of the three synthesized structures, are summarized in Fig{~\ref{fig:compare}}. When we compare GMM and mGMM, mGMM was found to be generally better than GMM. However, mGMM performances fluctuate largely as $R$, which is the size of $(\mu,\sigma)$ pairs in each algorithm, changes. It is worth noting that mGMM performed the best for the structure $S_1$, when the number of mGMM's mixture component was 1. However, mGMM suffers as R increases, implying that the increase of expressive power deteriorates mGMM performance. Given a real-world dataset, the exact number of mGMM's mixture components is unknown. In contrast, TGP is observed to have increasing performance in RMSE as the $R$ increases. Moreover, it is observed to have little fluctuations.

The inferior performance of the original GMM can be due to the lack of incorporating the sequential information of a trajectory and the permutation ambiguity in the Gaussian mixtures. In contrast, mGMM decomposes the approximation of the mean and co-variance into two steps. It first approximates the mean using K-mean++ algorithm, which incorporates the sequential information available in the trajectories. Using the K centroids, the co-variance for data that belongs to each centroid is then computed using singular value decomposition. This modification allows mGMM to outperform GMM for structures $S_1$ and $S_3$. However, this mGMM modification prevents the joint optimization of mean and co-variance simultaneously. The proposed TGP model, by contrast, naturally respects the sequential information of trajectory data and allows joint optimization of mean and co-variance functions of GP. In addition, TGP can handle trajectory with variable numbers of data points and variable intervals directly in both the training and inference phases, eliminating the need of interpolation that can distort the trajectory data.

According to the empirical results from the synthesized data, TGP demonstrates its capability in handling structures that undergo a variety of continuous deformations (e.g., extrusion, elongation, bend, twist and tension) to a Gaussian, a 5-petalled rose and Gaussian mixtures. This experiment and its findings provide evidence that TGP can better reconstruct the probabilistic structures given a dense point cloud compared to GMM and mGMM.

\subsection{A-SMGCS Dataset}
\begin{figure}[ht!]
    \centering
    \begin{subfigure}[b]{\textwidth}
        \centering
        \includegraphics[trim={0.5cm 0.5cm 0.5cm 0.5cm},clip,width=0.45\linewidth]{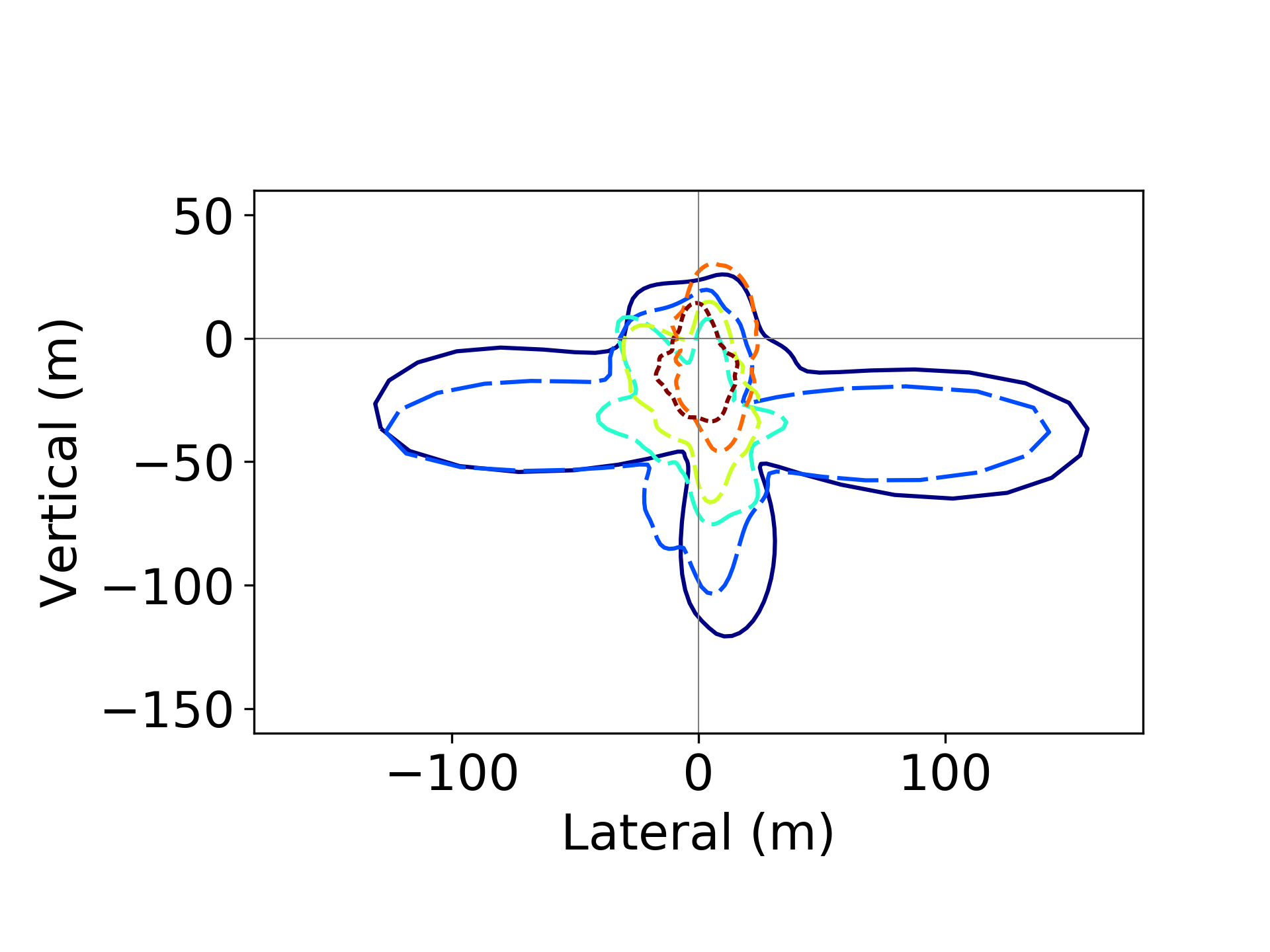}%
        \hspace{0.5cm}
        \includegraphics[trim={0.5cm 0.5cm 0.5cm 0.5cm},clip,width=0.45\linewidth]{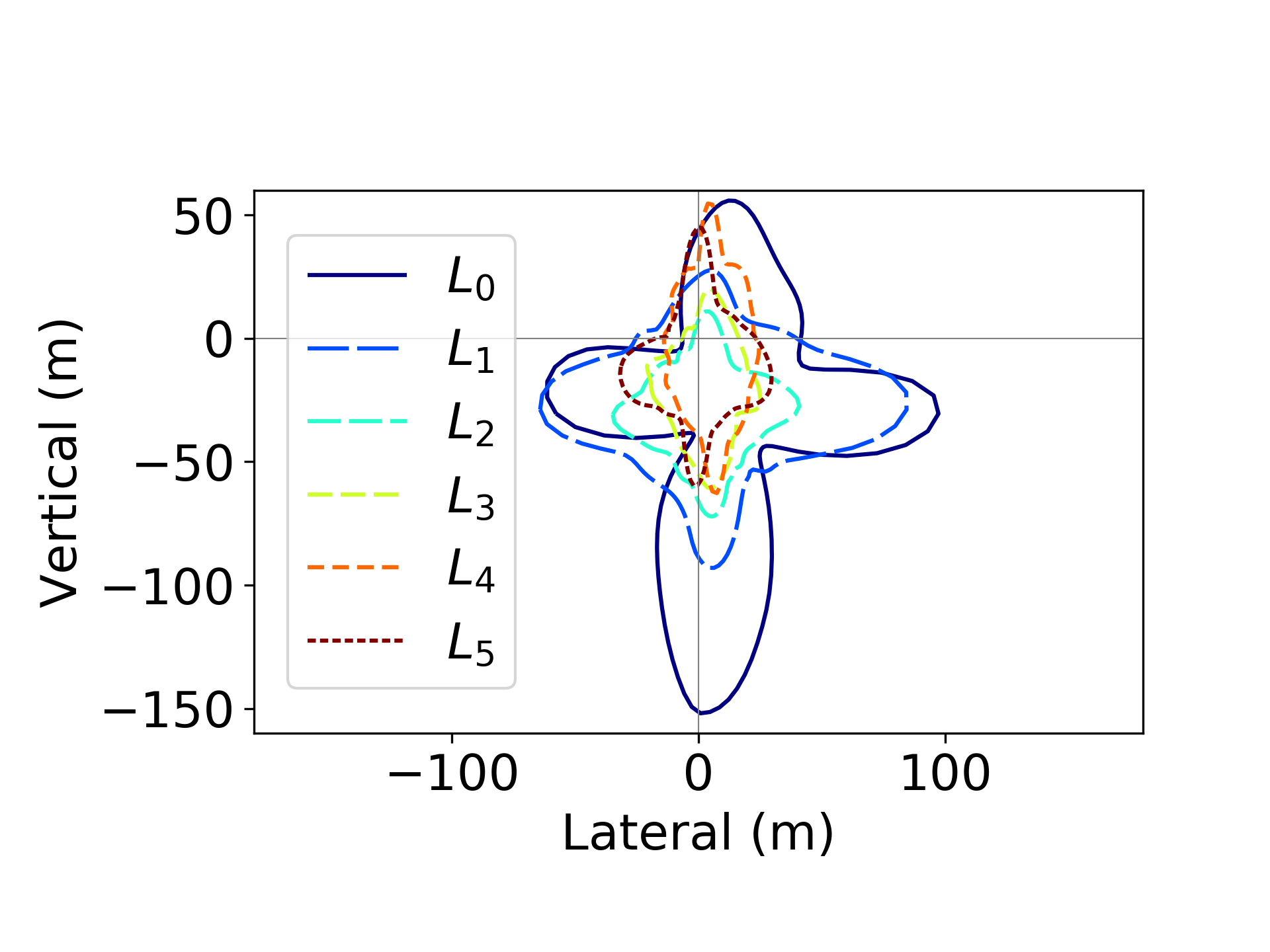}
    \end{subfigure}
    \caption{The cross-sectional view of probabilistic landing tunnel, for medium (left) and heavy (heavy) categories. $L_0$ to $L_5$ are uniformly distributed from AKIPO to ABVON.}
    \label{fig:crossview}
\end{figure}

\begin{figure}[ht!]
    \centering
    \includegraphics[width=0.9\linewidth]{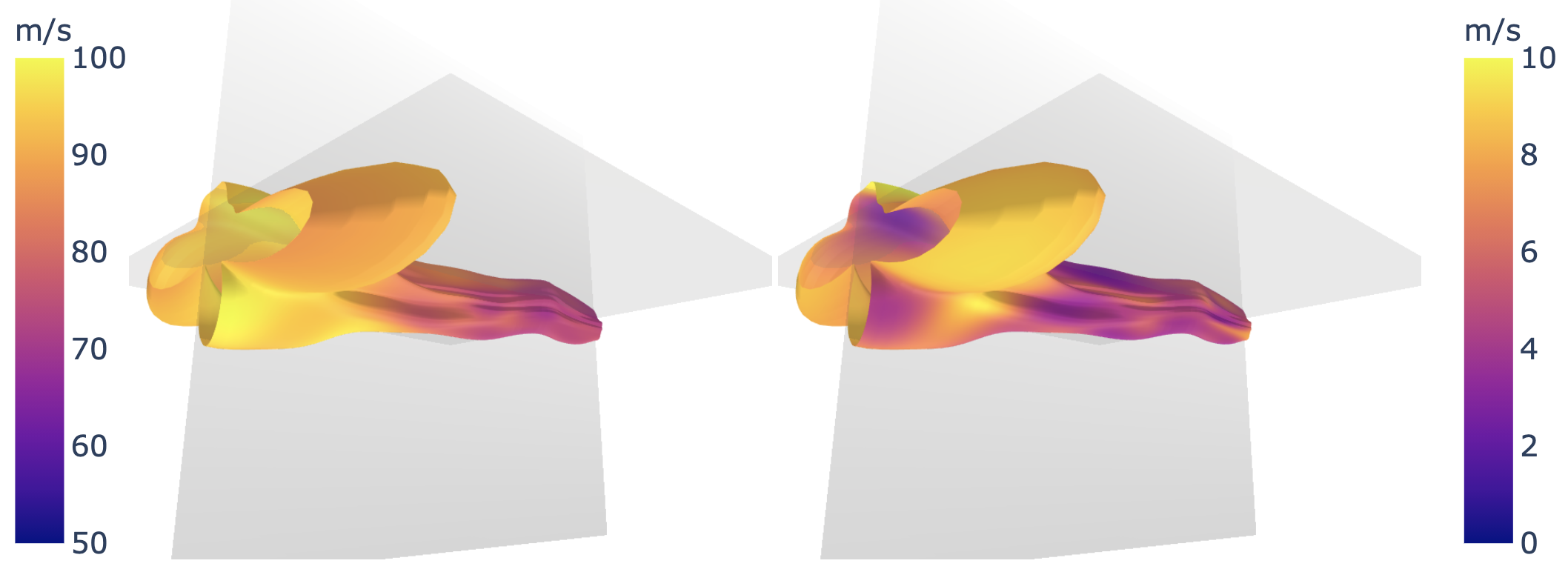}
    \caption{Reconstructed probabilistic landing tunnel using $TGP_1$ from medium category. Mean (left) and std (right) $S_x$ defined on $TGP_2$ are color-coded.}
    \label{fig:tunnel4DMx}
\end{figure}

\begin{figure}[ht!]
    \centering
    \includegraphics[width=0.9\linewidth]{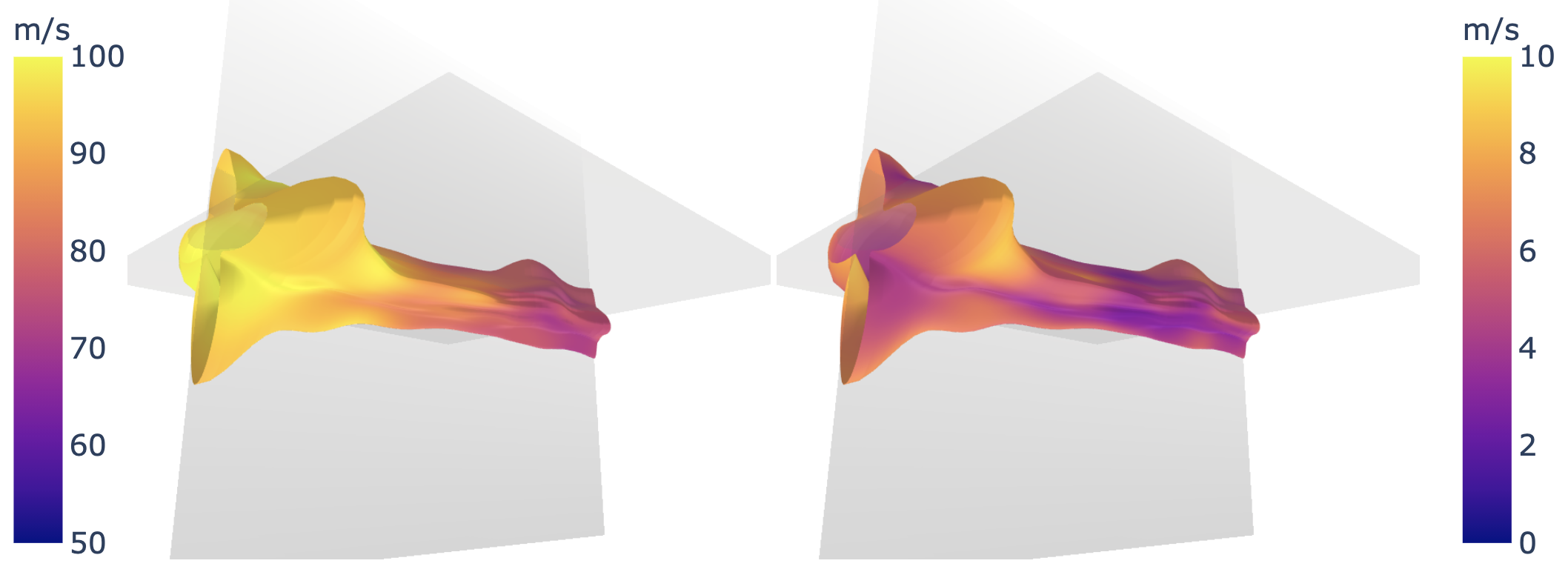}
    \caption{Reconstructed probabilistic landing tunnel using $TGP_1$ from heavy category. Mean (left) and std (right) $S_x$ defined on $TGP_2$ are color-coded.}
    \label{fig:tunnel4DHx}
\end{figure}

\begin{figure}[htb]
    \centering
    \begin{subfigure}[b]{\textwidth}
        \centering
        \includegraphics[width=0.45\linewidth]{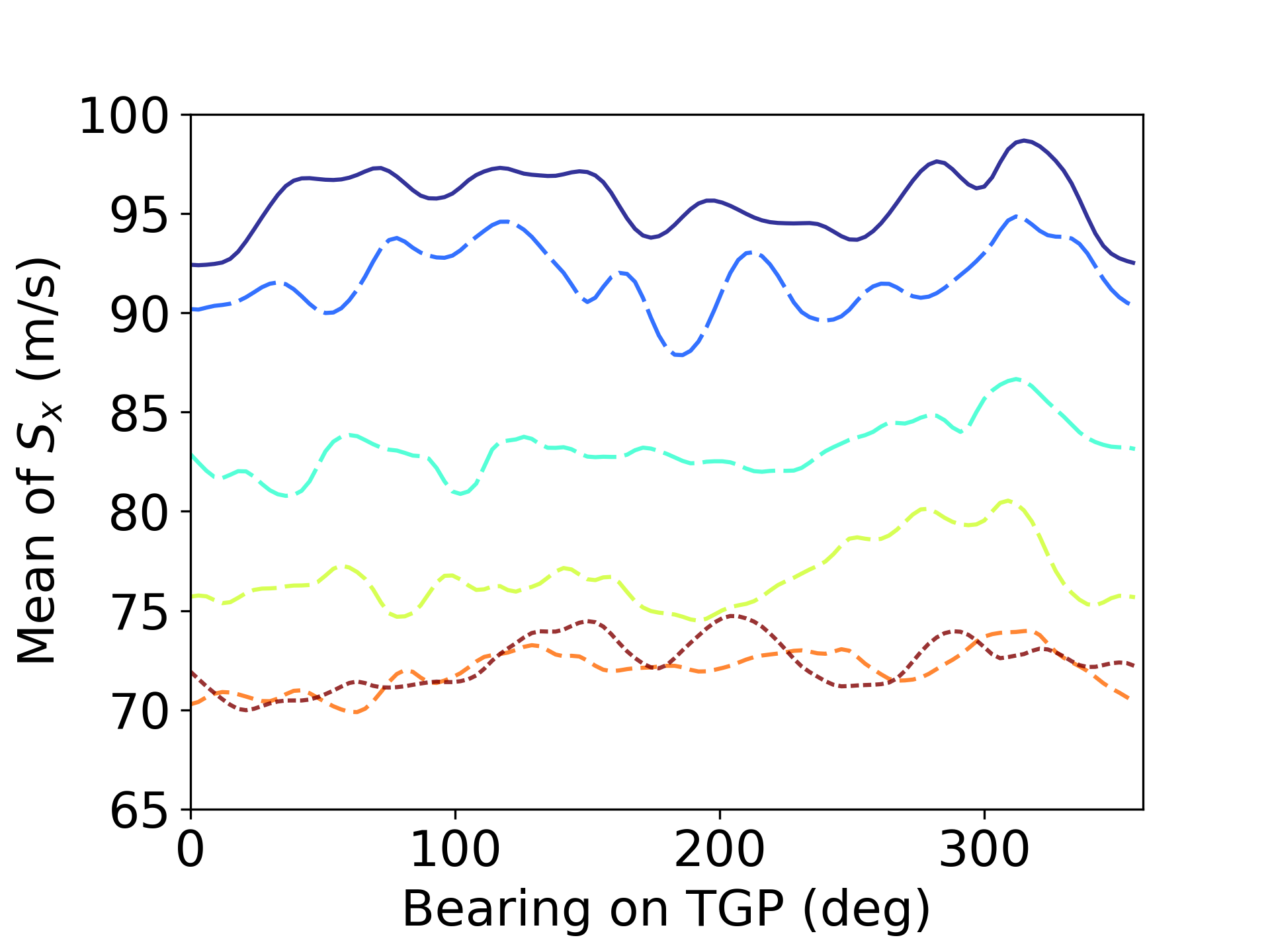}%
        \includegraphics[width=0.45\linewidth]{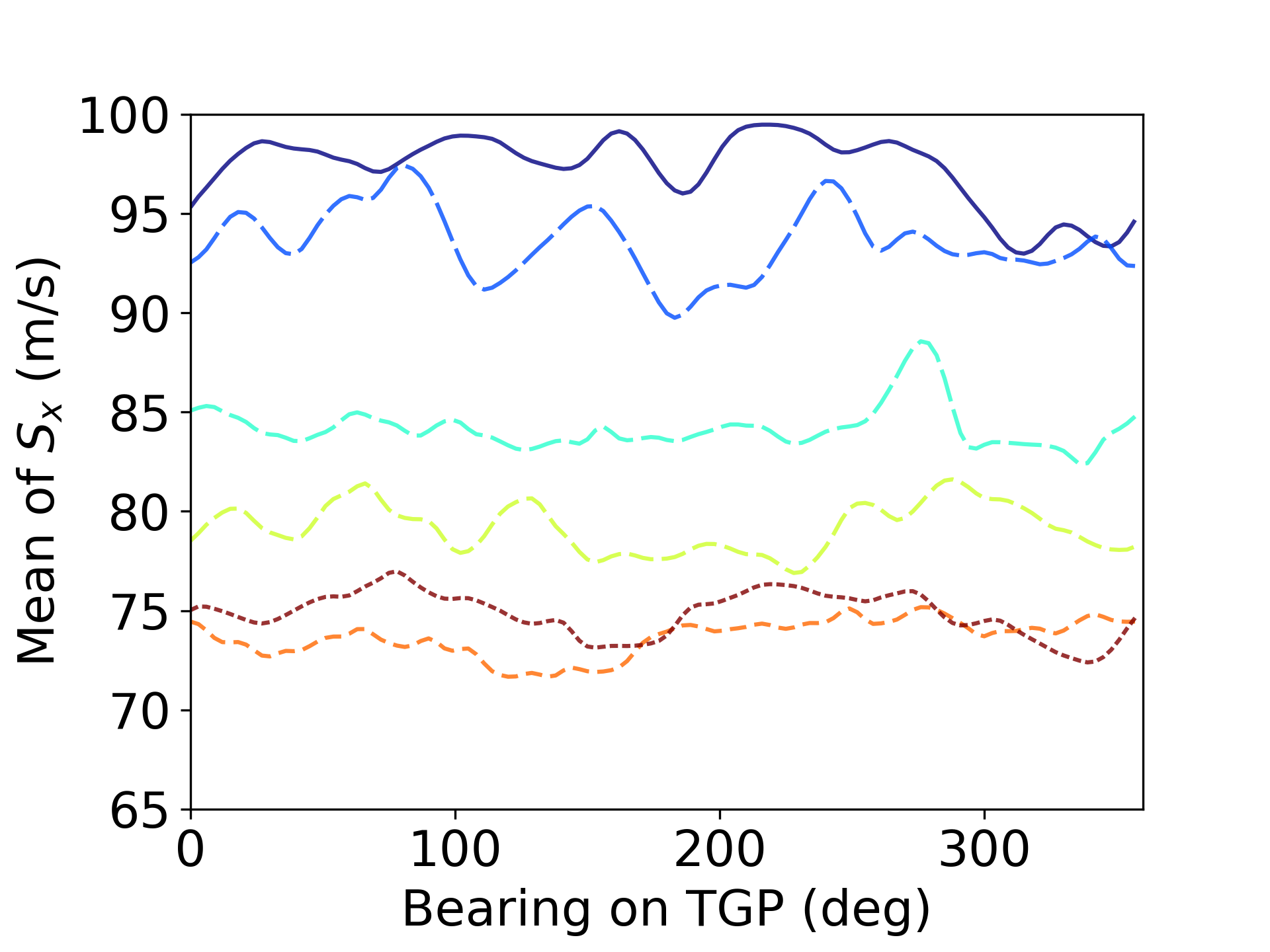}
        \caption{Mean of $S_x$ along landing glide slope, for medium (left) and heavy (right) categories.}
    \end{subfigure}
    \vskip 0.0001 \baselineskip
    \begin{subfigure}[b]{\textwidth}
        \centering
        \includegraphics[width=0.45\linewidth]{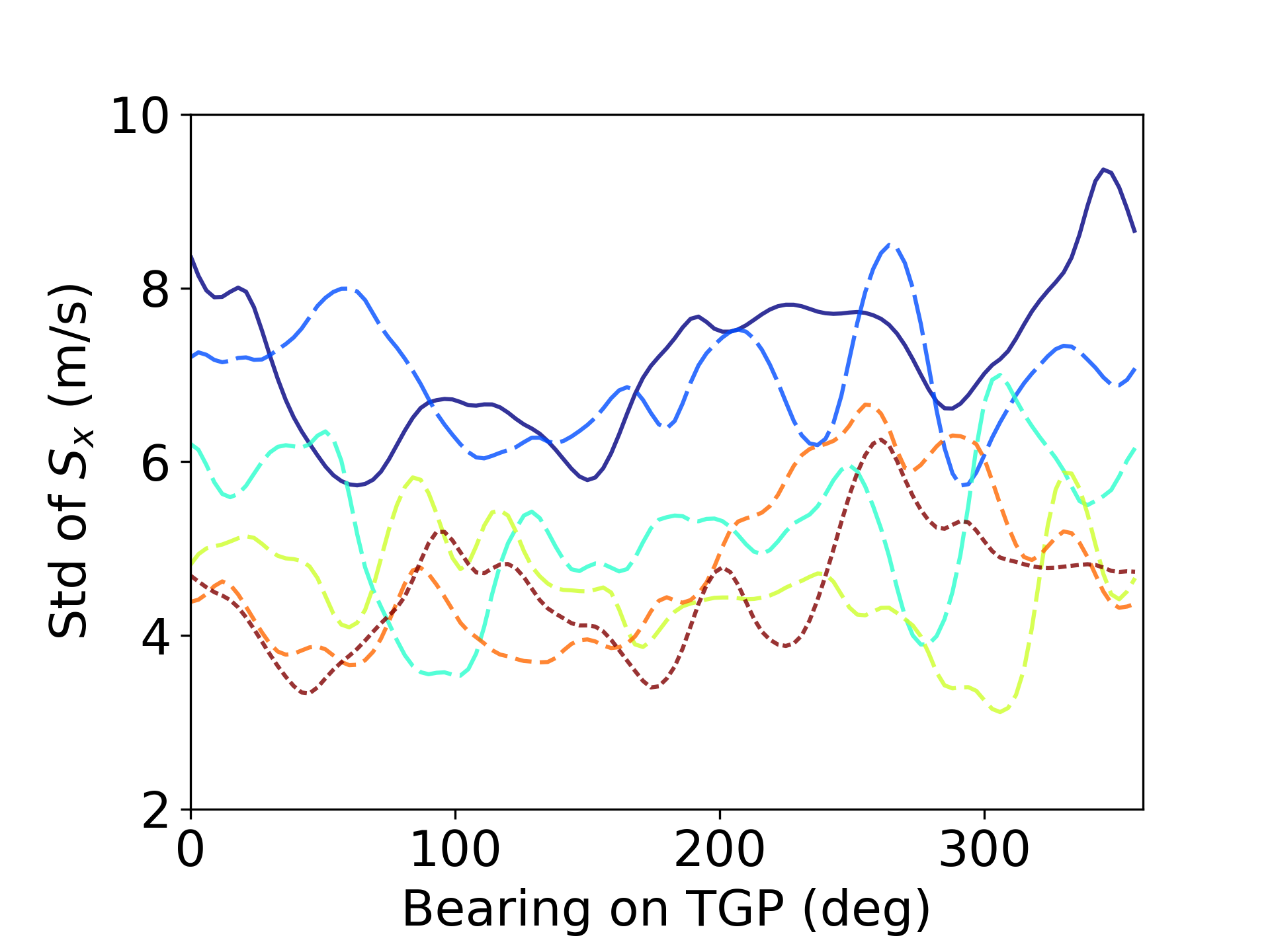}%
        \includegraphics[width=0.45\linewidth]{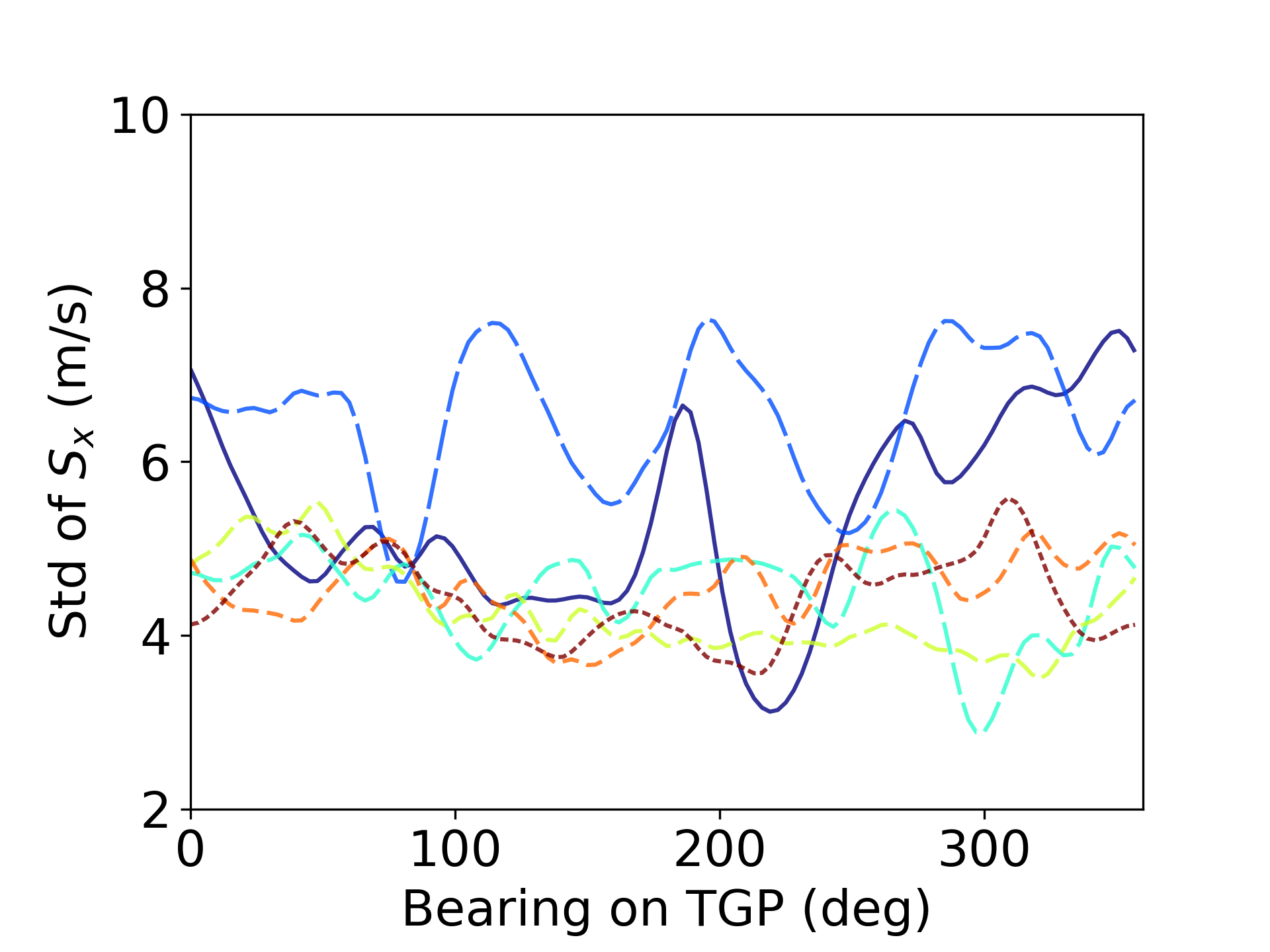}
        \caption{Std of $S_x$ along landing glide slope, for medium (left) and heavy (right) categories.}
    \end{subfigure}
    \caption{Landing parameter $S_x$. The legend can be found in Fig.~\ref{fig:crossview}.}
    \label{fig:tunnel2DSx}
\end{figure}

\begin{figure*}
    \centering
    \begin{subfigure}[b]{1\textwidth}
        \centering
        \includegraphics[trim={0cm 0cm 0cm 0cm},clip, width=0.9\textwidth,height=5cm]{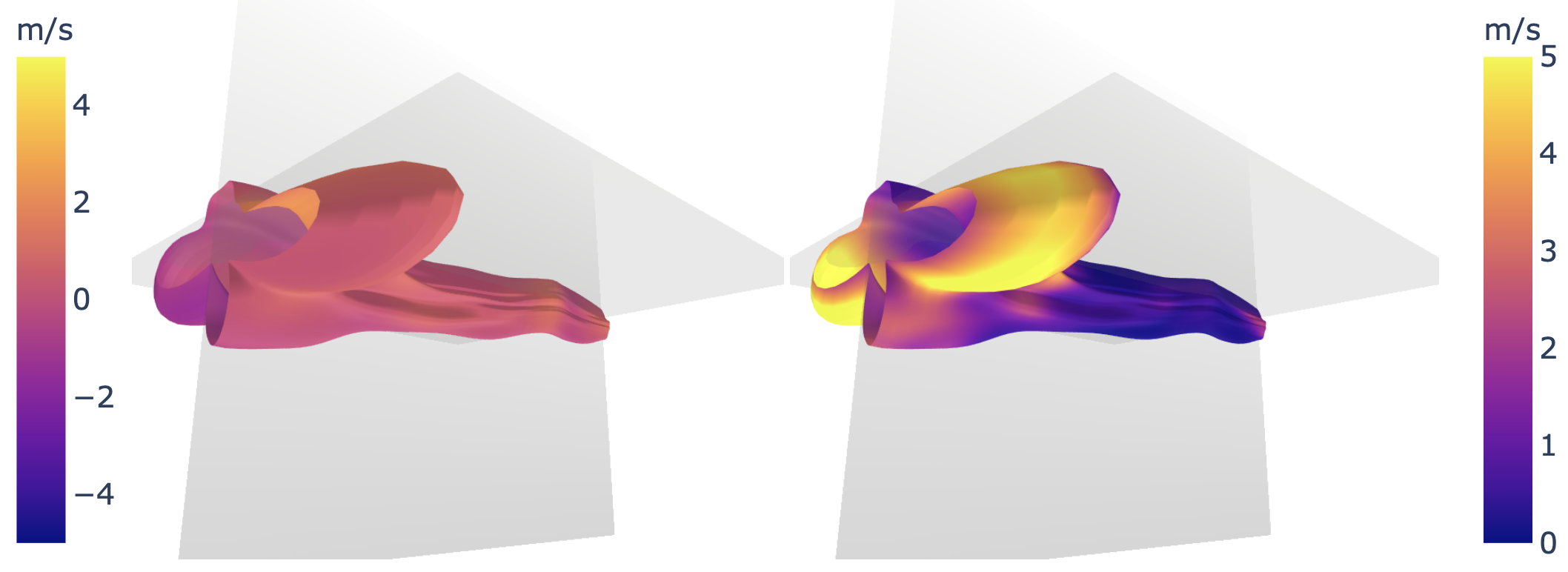}
    \end{subfigure}
    \vskip 0.0001 \baselineskip
    \begin{subfigure}[b]{0.45\textwidth}   
        \centering 
        \includegraphics[width=\textwidth,height=5cm]{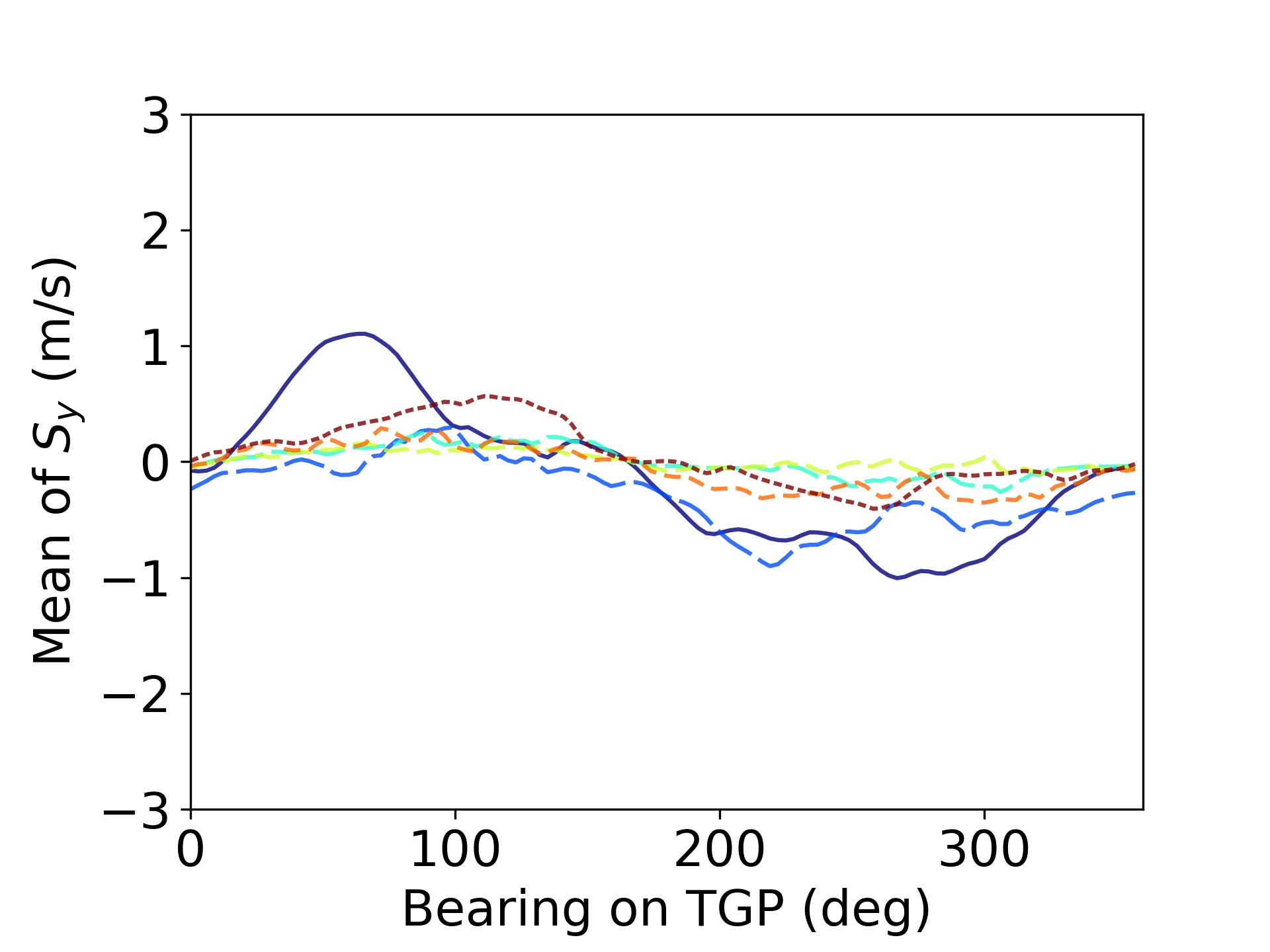}
    \end{subfigure}
    \begin{subfigure}[b]{0.45\textwidth}   
        \centering 
        \includegraphics[width=\textwidth,height=5cm]{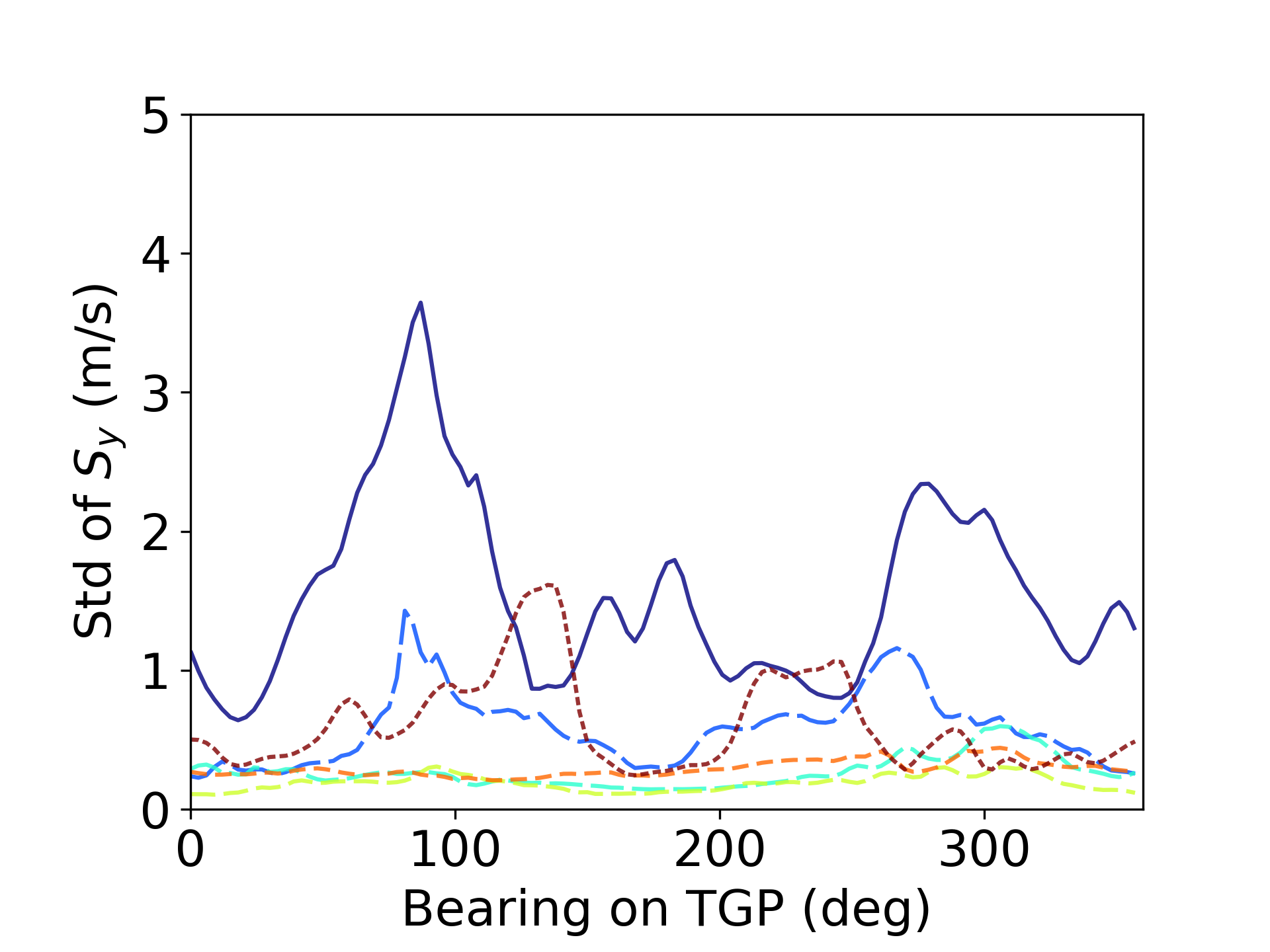}
    \end{subfigure}
    \caption[The average and standard deviation of critical parameters ]
    {\small The color-coded mean (left) and std (right) of $S_y$ on the reconstructed probabilistic landing tunnel (top) and the unfolded view (bottom). The legend can be found in Fig.~\ref{fig:crossview}.} 
    \label{fig:tunnel4DMy}
\end{figure*}

\begin{figure*}
    \centering
    \begin{subfigure}[b]{1\textwidth}
        \centering
        \includegraphics[trim={0cm 0cm 0cm 0cm},clip, width=0.9\textwidth,height=5cm]{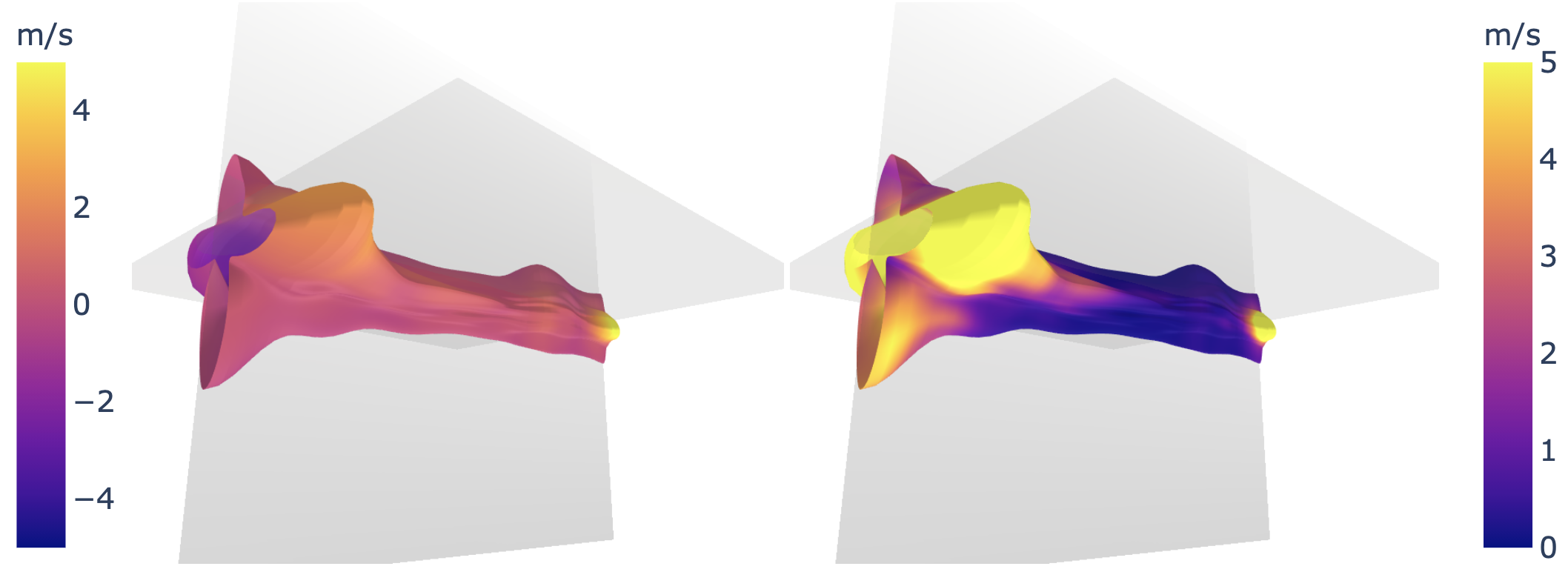}
    \end{subfigure}
    \vskip 0.0001 \baselineskip
    \begin{subfigure}[b]{0.45\textwidth}   
        \centering 
        \includegraphics[width=\textwidth,height=5cm]{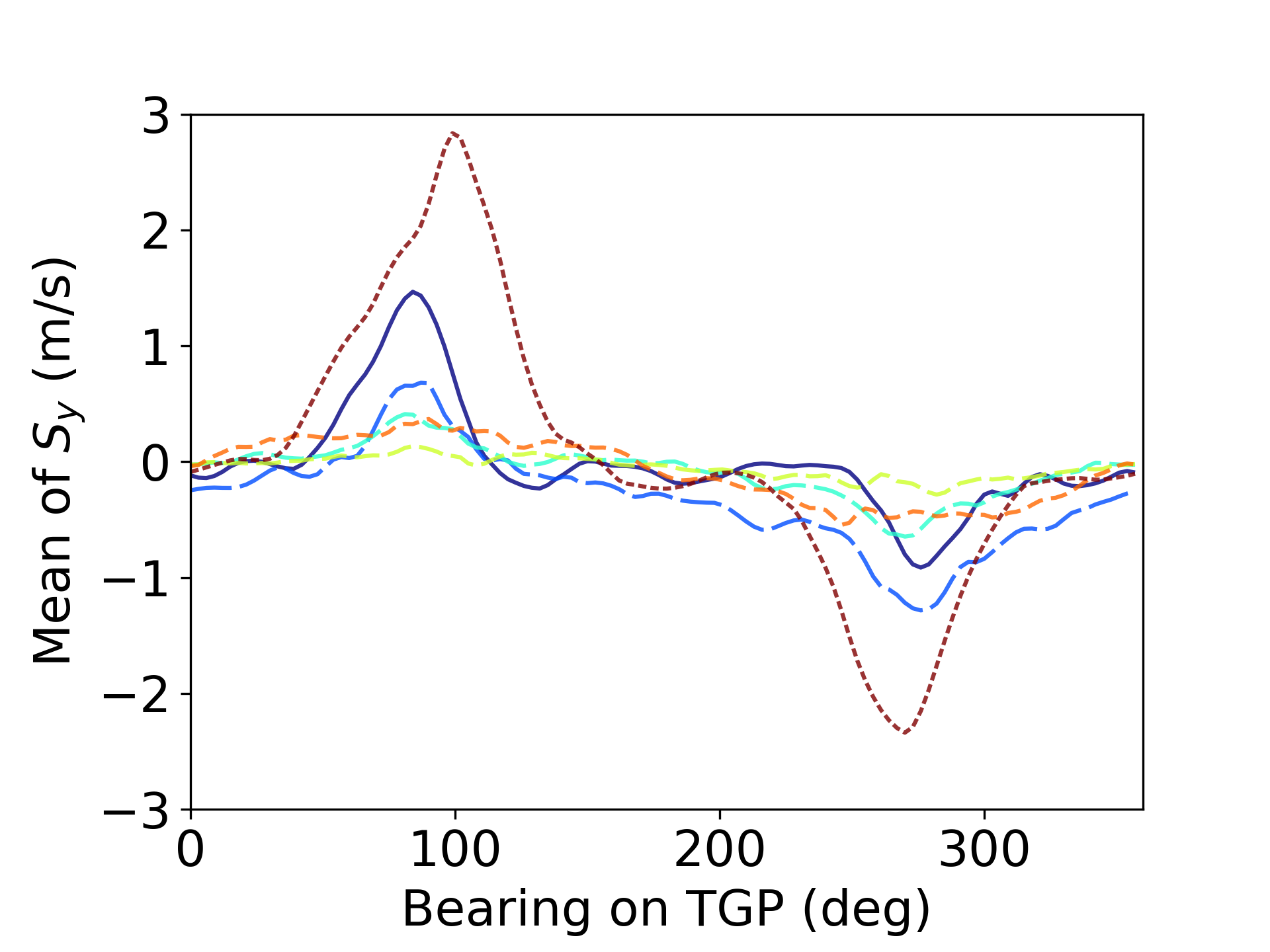}
    \end{subfigure}
    \begin{subfigure}[b]{0.45\textwidth}   
        \centering 
        \includegraphics[width=\textwidth,height=5cm]{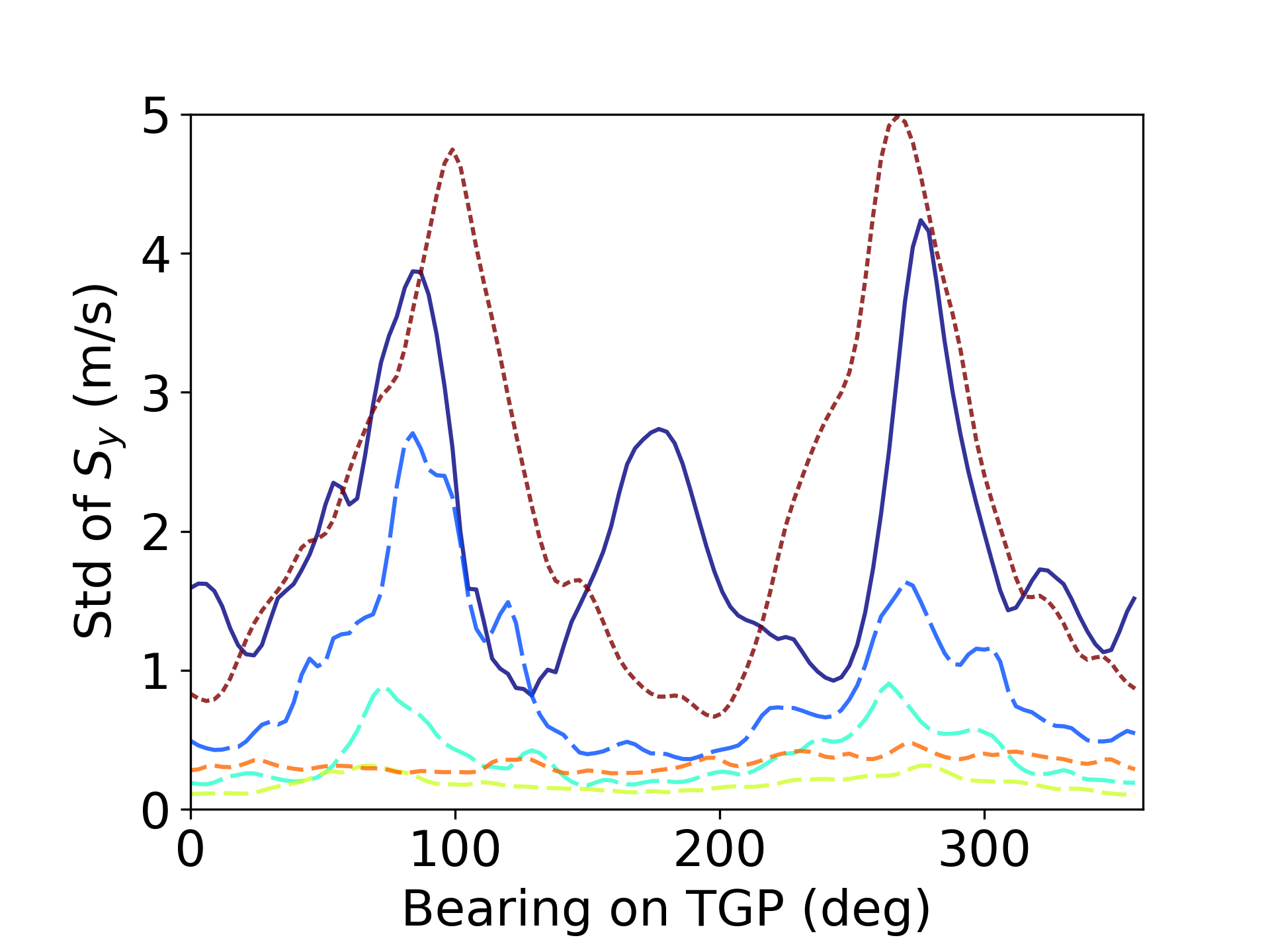}
    \end{subfigure}
    \caption[ The average and standard deviation of critical parameters ]
    {\small The color-coded mean (left) and std (right) of $S_y$ on the reconstructed probabilistic landing tunnel (top) and the unfolded view (bottom). The legend can be found in Fig.~\ref{fig:crossview}.} 
    \label{fig:tunnel4DHy}
\end{figure*}

\begin{figure*}
    \centering
    \begin{subfigure}[b]{1\textwidth}
        \centering
        \includegraphics[trim={0cm 0cm 0cm 0cm},clip, width=0.9\textwidth,height=5cm]{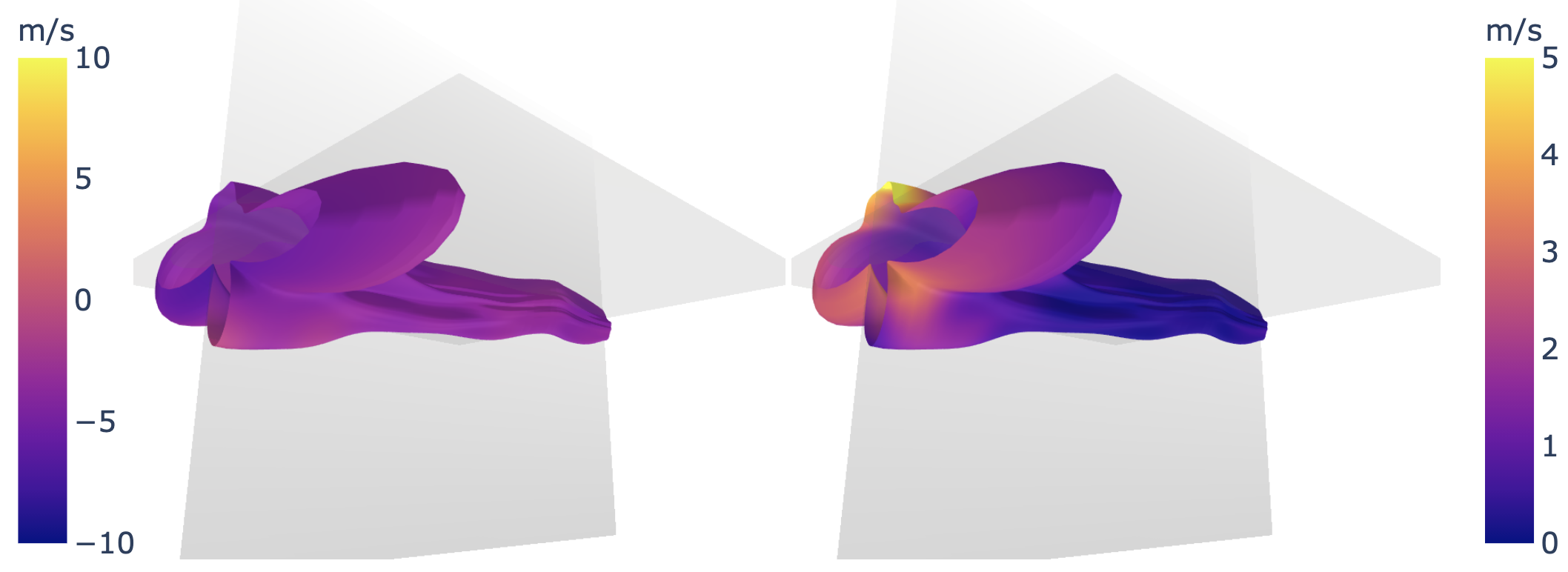}
    \end{subfigure}
    \vskip 0.0001 \baselineskip
    \begin{subfigure}[b]{0.45\textwidth}   
        \centering 
        \includegraphics[width=\textwidth,height=5cm]{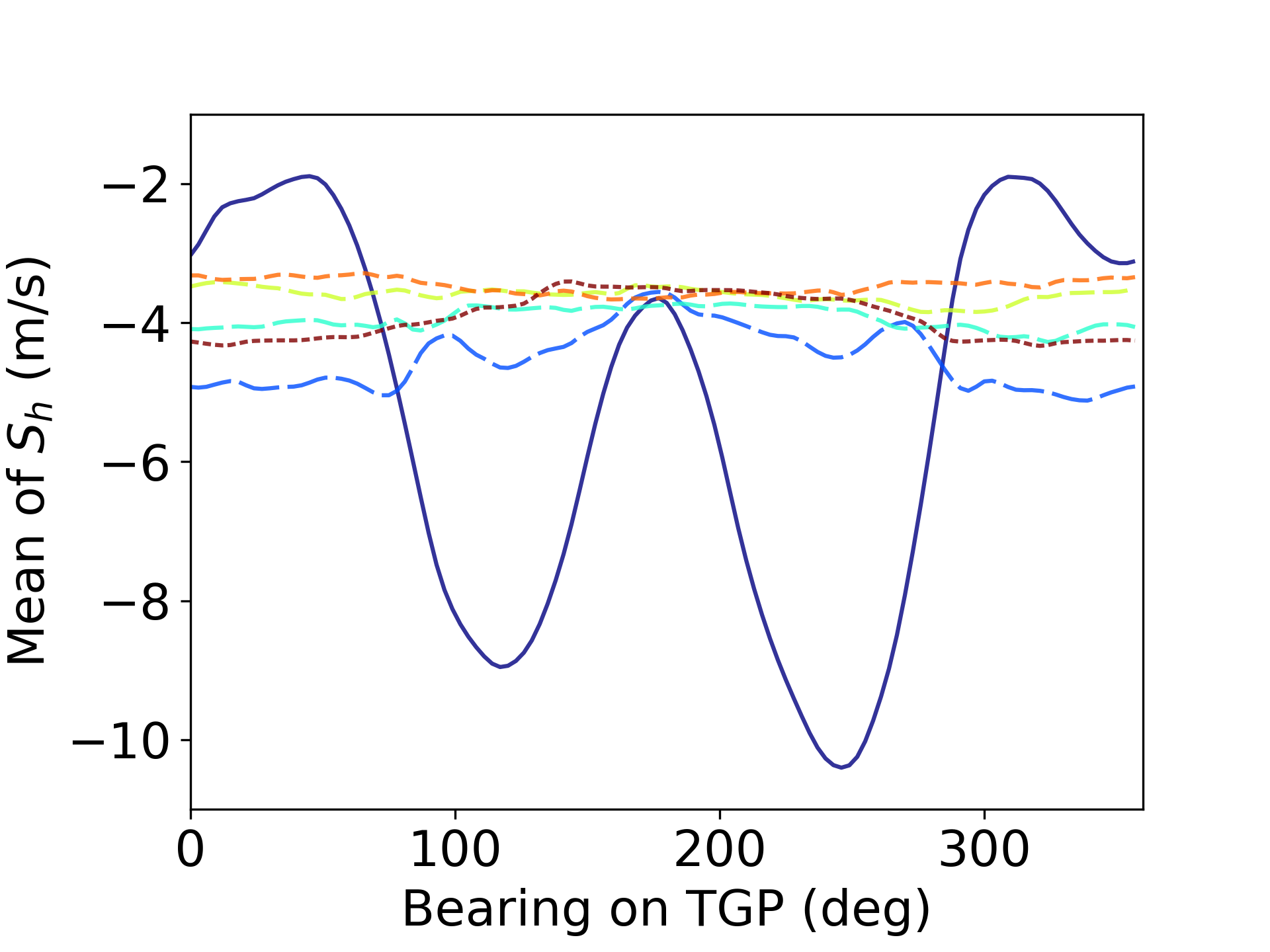}
    \end{subfigure}
    \begin{subfigure}[b]{0.45\textwidth}   
        \centering 
        \includegraphics[width=\textwidth,height=5cm]{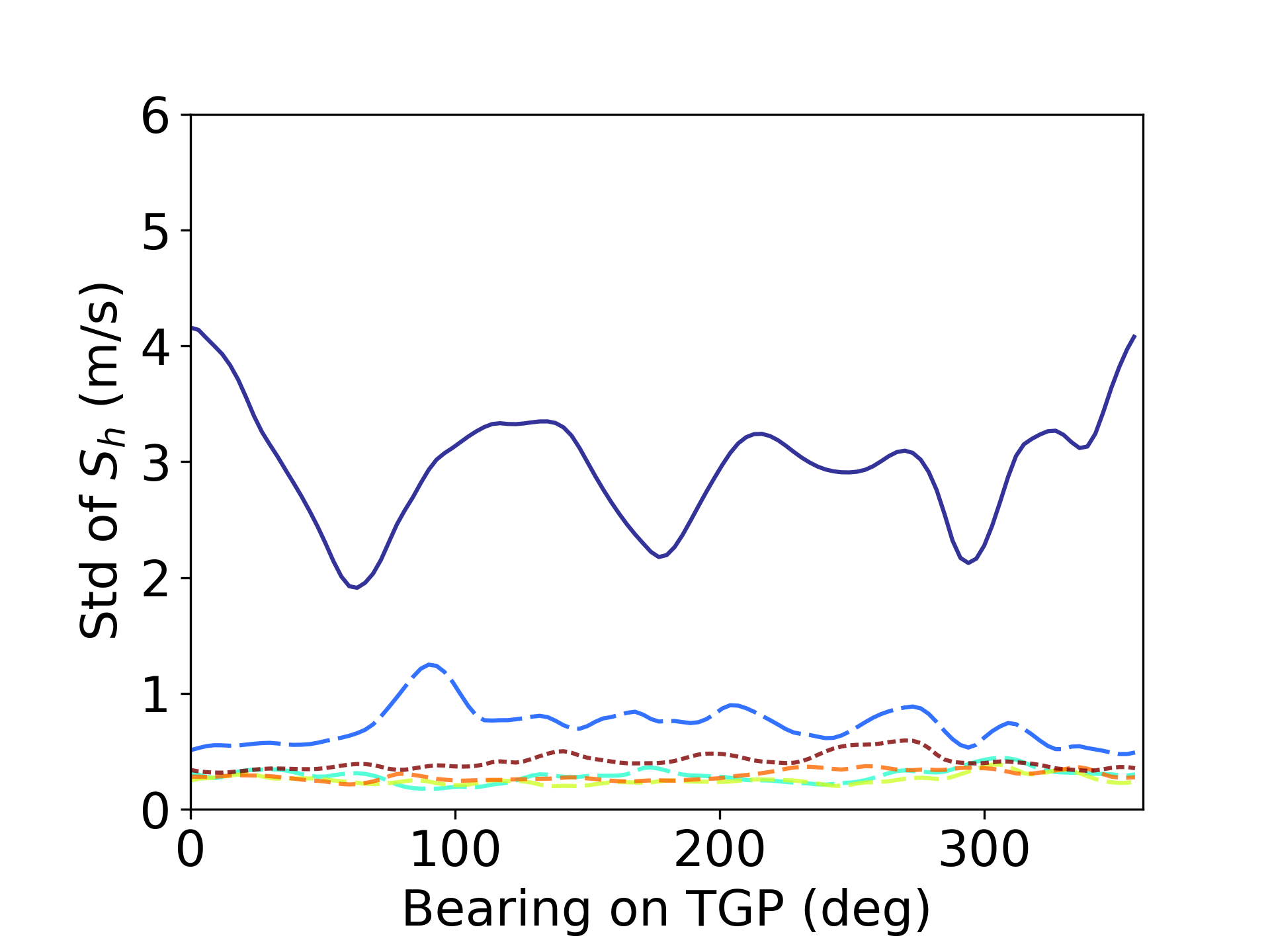}
    \end{subfigure}
    \caption[The average and standard deviation of critical parameters ]
    {\small The color-coded mean (left) and std (right) of $S_h$ on the reconstructed probabilistic landing tunnel (top) and the unfolded view (bottom). The legend can be found in Fig.~\ref{fig:crossview}.} 
    \label{fig:tunnel4DMh}
\end{figure*}

\begin{figure*}
    \centering
    \begin{subfigure}[b]{1\textwidth}
        \centering
        \includegraphics[trim={0cm 0cm 0cm 0cm},clip, width=0.9\textwidth,height=5cm]{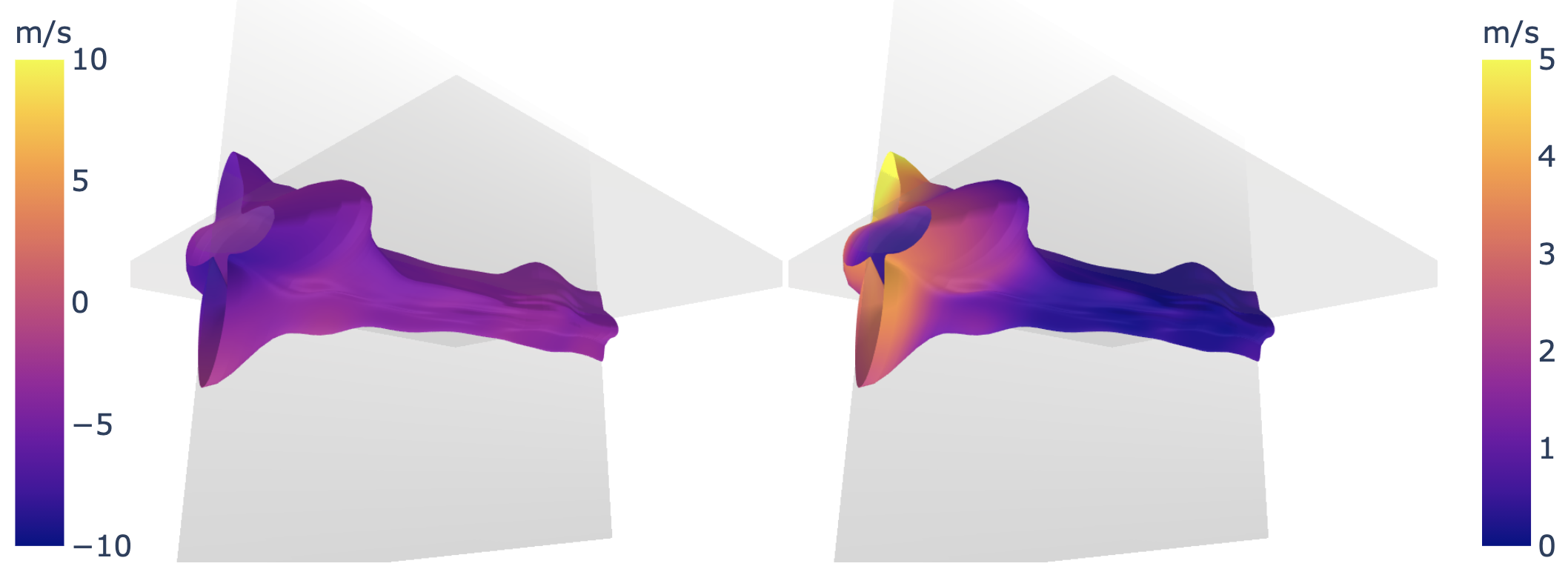}
    \end{subfigure}
    \vskip 0.0001 \baselineskip
    \begin{subfigure}[b]{0.45\textwidth}   
        \centering 
        \includegraphics[width=\textwidth,height=5cm]{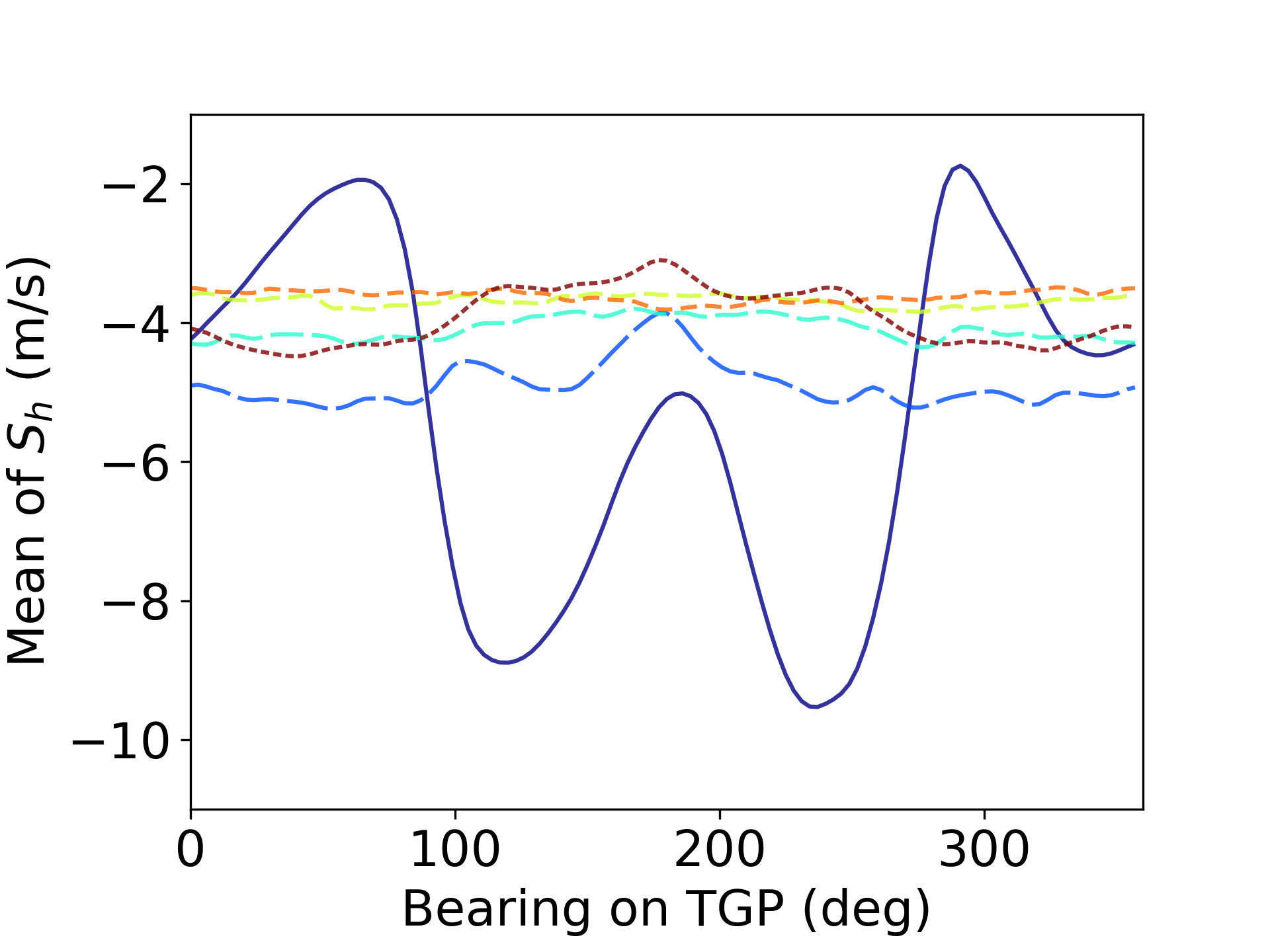}
    \end{subfigure}
    \begin{subfigure}[b]{0.45\textwidth}   
        \centering 
        \includegraphics[width=\textwidth,height=5cm]{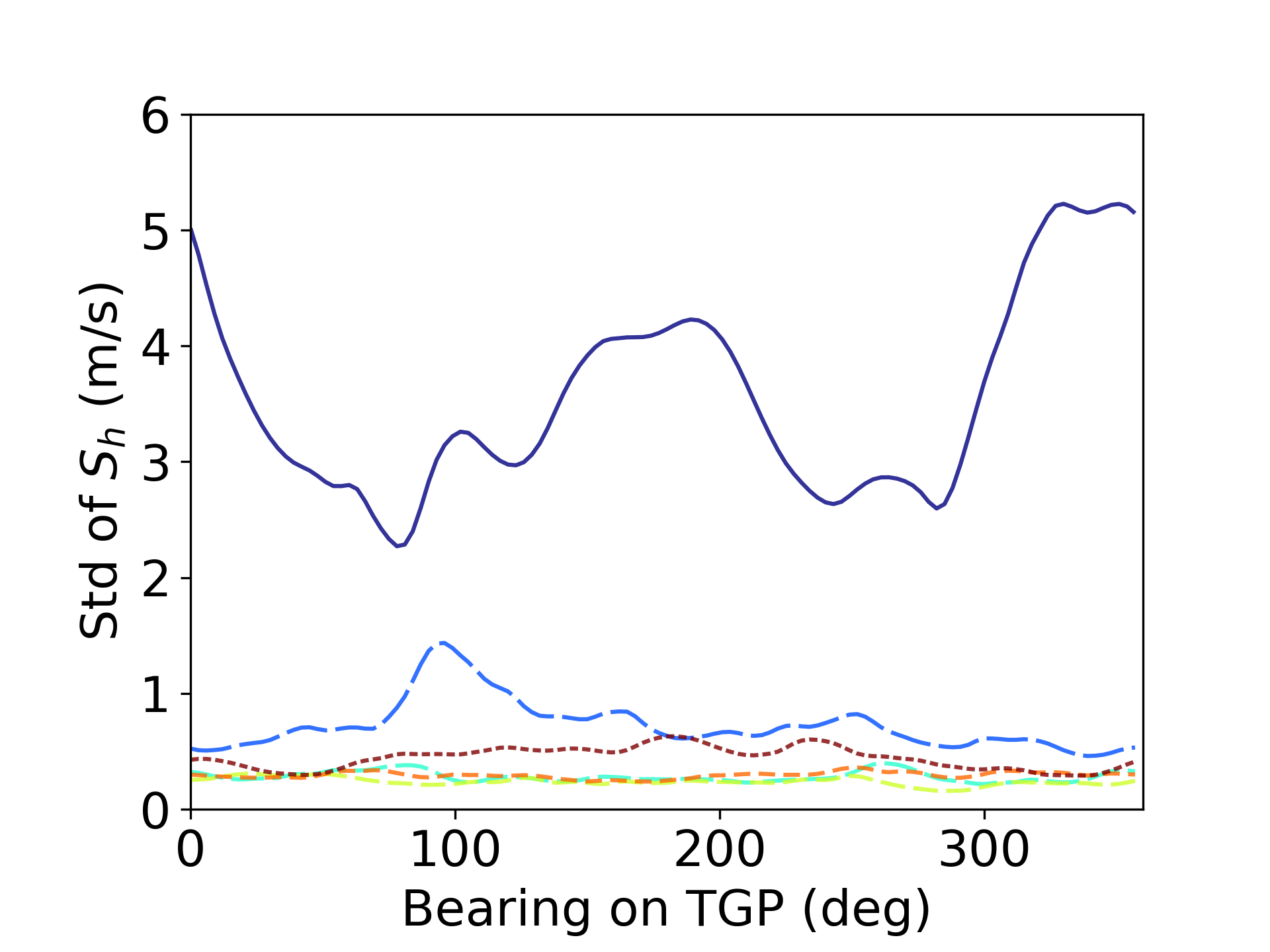}
    \end{subfigure}
    \caption[ The average and standard deviation of critical parameters ]
    {\small The color-coded mean (left) and std (right) of $S_h$ on the reconstructed probabilistic landing tunnel (top) and the unfolded view (bottom). The legend can be found in Fig.~\ref{fig:crossview}.} 
    \label{fig:tunnel4DHh}
\end{figure*}

This section examines and analyzes TGP model 1 for reconstructing the probabilistic approach and landing tunnel and TGP model 2 for approach and landing parameters (i.e., velocity). Fig.~\ref{fig:crossview} illustrates the projected view of the tunnel when one is looking from the cockpit uniformly along the glide slope, while Fig.~\ref{fig:tunnel4DMx} $\&$~\ref{fig:tunnel4DHx} depict the 4D reconstruction for medium and heavy category respectively. These probabilistic tunnels are generated using two $\sigma_*$ bounds. On the tunnels in Fig.~\ref{fig:tunnel4DMx} $\&$~\ref{fig:tunnel4DHx}, the mean and std of inferred speed $S_x$ are color-coded. Similarly, the mean and std of inferred speed $S_y$ and $S_h$ are provided in Fig.~\ref{fig:tunnel4DMy}-~\ref{fig:tunnel4DHh}. Also, we show the unfolded view of the mean and std of inferred speed $S_x$, $S_y$ and $S_h$ in Fig.~\ref{fig:tunnel2DSx} -~\ref{fig:tunnel4DHh}.

In Fig.~\ref{fig:crossview},~\ref{fig:tunnel4DMx} $\&$,~\ref{fig:tunnel4DHx}, we observe that aircraft aligns with the ILS guidance and the space utilization is generally decreasing for both medium and heavy categories. Nonetheless, we can visually distinguish between the tunnels of heavy and medium categories. It is observed that the medium category occupies more space laterally while the heavy category occupies more space vertically, when the aircraft is closer to AKIPO (an intermediate fix on $3^0$ glide slope of ILS which is 7.8 Nautical miles from touch down of Runway 02L of Changi Airport, Singapore). Since the inertia of medium category aircraft is lesser than the heavy category, center-line alignment were performed near AKIPO. Hence, the bounds are bigger for medium category aircraft. Heavy category aircraft appears to be more structured while carrying out precision approach landing, which is consistent with the fact that they require a longer distance from touch down to be properly aligned with runway center-line and on the glide slope.

It is interesting to note that the majority of aircraft appear below the glide slope, as aircraft has little energy excess to dissipate. Based on the go-around reports, many aircraft flying above the glide slope ended up in an unstable approach configuration. The heavy category appears to have a high probability of being above the glide slope. Laterally and vertically, both tunnels are not exactly symmetrical. Hence, our TGP models can reveal these intrinsic structures. 

In Fig.~\ref{fig:tunnel4DMx},~\ref{fig:tunnel4DHx} and~\ref{fig:tunnel2DSx}, we found that the mean and std of $S_x$ are generally decreasing along the glide slope, which are consistent with aircraft landing deceleration characteristics. Fig.~\ref{fig:tunnel4DMy} and~\ref{fig:tunnel4DHy} show that the mean and std of $S_y$ are generally converging to zero along the glide slope, providing insights into localizer alignment. Greater variations in the mean and std of speed are found when aircraft align with the center-line laterally. These variations converge when aircraft get closer to ABVON (A fix designated as missed approach point at or before which an aircraft must initiate go-around if aircraft safety is in jeopardy).

Fig.~\ref{fig:tunnel4DMh} and~\ref{fig:tunnel4DHh} depict the descent rate $S_h$. As aircraft travels from the intermediate fix AKIPO to missed approach point ABVON, the mean of $S_h$ is converging to a speed below 5.08 m/s (1000ft/min, stabilized approach criterion) and the std of $S_h$ are generally converging to a value close to zero, along the glide slope. Greater variations in descent rate were observed near AKIPO which is in line with the data-set as we observed that aircraft are being vectored by ATC to make final alignment with ILS around fix AKIPO. Sometimes due to inertia and height, aircraft perform steeper descent to capture glide slope from above and center-line alignment simultaneously. 

To share more intriguing findings, we also provide additional analyses online. Our interactive visualizations are available at \url{https://simkuangoh.github.io/TunnelGP/}. The experiment was ran on Intel® Core™ i9-9900X CPU @ 3.50GHz with 32 GB RAM and Nvidia RTX2080Ti.

\subsection{Other applications of TGP}
\begin{figure}[ht!]
    \centering
    \includegraphics[width=\linewidth]{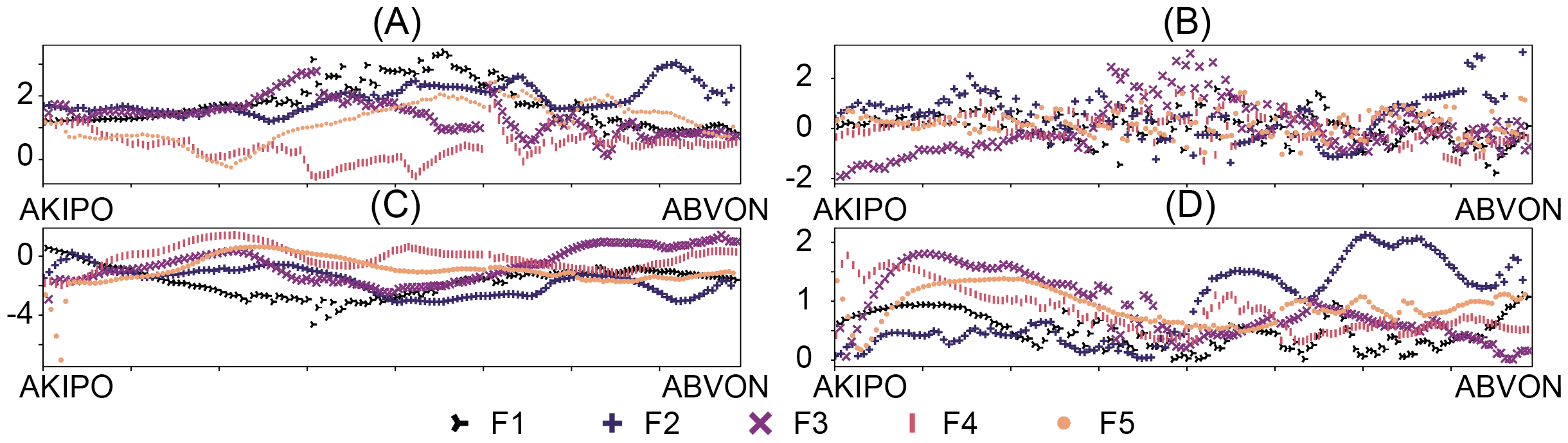}
    \caption{Streaming of data from five flights with identities F1-F5 into TGP models. Panel (A), (B), (C), and (D) illustrate the $Z$-scores, defined in equation~\ref{eq:Z}, for $S_x$, $S_y$, $S_h$, and $P$, respectively.}
    \label{fig:app}
\end{figure}

While this paper focuses on applying TGP for the probabilistic characterization and learning of approach and landing parameters, there are a number of other potential applications. The potential emerges from the fact that TGP explicitly models $(\mu,\sigma)$ in equation{~\ref{eqn:gp}}. Firstly, given $(\mu,\sigma)$, we can sample the trajectories that are similar to historical data. The operational uses of these probabilistic trajectory models have been discussed in{~\cite{barratt2018learning}}, which include assessment of the safety and operational performance of new technologies and procedures in a simulated environment. Secondly, the prediction interval of the future observation of approach and landing parameters can be defined for $[l,u]$, as follows:

\begin{equation}
\gamma = P(l<X<u)=P(\frac{l-\mu}{\sigma}<\frac{X-\mu}{\sigma}<\frac{u-\mu}{\sigma})<P(\frac{l-\mu}{\sigma}<Z<\frac{u-\mu}{\sigma}) 
\label{eq:Z}
\end{equation}

\noindent where $\mu$ and $\sigma$ are defined in equation{~\ref{eqn:gp}} and can be obtained from the trained TGP models. $Z$ is the standard score. Five approach and landing trajectories are streamed to TGP models and the associated $Z$-scores are illustrated in Fig.~\ref{fig:app}. The scores indicate the sign and size of the deviations in approach and landing parameters. 

Equation~\ref{eq:Z} provides the probability $\gamma$ that an approach parameter falls within $[l,u]$. These values can be adjusted flexibly depending on the desired sensitivity and specificity trade-off along the glide slope. This property facilitates automated conformance monitoring. ATCO and aircrew will only be alerted if the violation of these parameters are detected. Thirdly, it allows generation of aircraft approach and landing parameters that have anomalous measurements in a simulated environment, to assess other procedures and technologies. For example, a stabilized approach can be made unstable by modifying the approach parameters in such a way the $Z$ of approach and landing parameter in equation{~\ref{eq:Z}} gradually violates the desired range. Fourthly, the root cause of go-around has been reported to be undetectable from surveillance data in{~\cite{figuet2020predicting}}, which highlighted an important open issue in air traffic management. We will extend the TGP model to investigate the issue.

\section{Conclusion}\label{conclusion}

In this paper, we propose two variants of TGP models that draw inspiration from SVGP and polar GP. TGP addresses some of the limitations of SVGP and polar GP, but inherits their strengths, which can learn from a large amount of data in cylindrical coordinate, providing the probabilistic description of trajectories and their associated flight's parameters. Three synthesized datasets are generated for qualitative and quantitative assessment.

We assess TGP using synthesized data and acquired A-SMGCS data. According to the results, TGP reconstructs the probability density functions given their respective dense point cloud and provides the probabilistic boundaries that encapsulate the data. The probabilistic boundaries provide continuous descriptions of the uncertainties in approach and landing dynamics. These findings will facilitate the analysis of landing and associated hazards. Moreover, it can provide an alternative supportive tool for periodical safety assessment to the air navigation service providers (ANSPs). 

The cross-sectional views of TGP provide us useful insights regarding aircraft parameter deviations (i.e., position and velocity). Moreover, these views are represented as if they are seen from the cockpit, which will enhance the situational awareness of ATCOs if presented to them. In real-time, aircraft landing dynamics and its adherence to stabilized approach criteria can be monitored by ATCOs. Any non-compliance will trigger a warning and controllers will be alerted.  As a result, they can advise aircrew about any non-adherence to stabilized approach criteria accordingly. Also, it will provide additional ground-based safety nets and will complement the existing ground proximity warning system (GPWS), which is an airborne safety net available to aircrew.

On the parallel runway approaches where safety requirement is very stringent~\cite{liang2018conflict}, this model can be utilized to carry out a periodical safety assessment and hazard analysis. Based on the historical data of parallel approaches and its safety assessment, future designs of different parallel approaches can be proposed by ANSPs. In addition, this model can be augmented in designing satellite-based approaches in the future. Moreover, collision risk between different runways and landing configurations can be assessed based on the probabilistic bounds derived from TGP. This paves the way for complex landing procedures such as parallel approach.

\section*{Acknowledgment}
This research is supported by the National Research Foundation, Singapore, and the Civil Aviation Authority of Singapore, under the Aviation Transformation Programme. Any opinions, findings and conclusions or recommendations expressed in this material are those of the author(s) and do not reflect the views of National Research Foundation, Singapore and the Civil Aviation Authority of Singapore.

\bibliography{sample}

\end{document}